\documentclass[10pt,twocolumn,letterpaper]{article}

\usepackage[pagenumbers]{arxiv_twocolumn} 

\usepackage{fontawesome5}

\definecolor{cvprblue}{rgb}{0.21,0.49,0.74}
\usepackage[pagebackref,breaklinks,colorlinks,allcolors=cvprblue]{hyperref}

\def\paperID{*****} 
\def\confName{CVPR}
\def\confYear{2026}

\definecolor{newgray}{gray}{0.4}

\usepackage[most]{tcolorbox}

\definecolor{headingpurple}{HTML}{7030A0}
\definecolor{BoxPurple}{HTML}{7040A0}
\definecolor{BorderPurple}{HTML}{60358A}
\definecolor{LightPurple}{HTML}{F8F5FC}

\hypersetup{
    colorlinks=true,
    linkcolor=headingpurple,
    citecolor=headingpurple!50,
    filecolor=headingpurple,      
    urlcolor=headingpurple,
    }

\newtcolorbox{questionbox}[1][Research Question]{
  enhanced,
  width=\linewidth,
  colback=LightPurple,
  colframe=BorderPurple,
  boxrule=0.7pt,
  arc=2mm,
  left=2.5mm,
  right=2.5mm,
  top=3mm,
  bottom=1.6mm,
  title={#1},
  fonttitle=\fontsize{9pt}{11pt}\selectfont\bfseries,
  coltitle=white,
  colbacktitle=BoxPurple,
  attach boxed title to top left={
    xshift=3mm,
    yshift=-\tcboxedtitleheight/2
  },
  boxed title style={
    boxrule=0pt,
    arc=1mm,
    left=2mm,
    right=2mm,
    top=0.2mm,
    bottom=0.2mm
  },
  before skip=10pt,
  after skip=8pt
}

\patchcmd{\cvprsection}
  {\raggedright}
  {\raggedright\color{headingpurple}}
  {}
  {\PackageWarning{heading-colors}{Failed to patch section}}

\patchcmd{\paragraph}
  {\bfseries}
  {\bfseries\color{headingpurple}}
  {}
  {\PackageWarning{heading-colors}{Failed to patch paragraph}}

\patchcmd{\cvprsubsection}
  {\raggedright}
  {\raggedright\color{headingpurple}}
  {}
  {\PackageWarning{heading-colors}{Failed to patch subsection}}

\patchcmd{\cvprsubsubsection}
  {\raggedright}
  {\raggedright\color{headingpurple}}
  {}
  {\PackageWarning{heading-colors}{Failed to patch subsubsection}}

\patchcmd{\abstract}
  {\centerline{\large\bf Abstract}}
  {\centerline{\large\bf\color{headingpurple} Abstract}}
  {}
  {\PackageWarning{heading-colors}{Failed to patch abstract}}

\title{PhysOPSD: Code as Privilege for Physically Plausible\\Video Generation}

\author{
\vspace{-2em}\\
\fontsize{11pt}{9.9pt}\selectfont
\textbf{Harold Haodong Chen$^{1,2,*}$,}
\textbf{Rui Peng$^{2,\dagger}$,}
\textbf{Lingteng Qiu$^{2}$,}\\
\textbf{Kejie Qiu$^{2}$,}
\textbf{Chao Xu$^{2}$,}
\textbf{Zilong Dong$^{2}$} \\
\\
\fontsize{10.4pt}{9.84pt}\selectfont
$^1$HKUST \& $^2$Alibaba Token Hub, Alibaba Group \\
\fontsize{10pt}{9.5pt}\selectfont
$^*$Work done at ATH@Alibaba Group, $^\dagger$Corresponding author\\
\vspace{-0.8em}
\\
\fontsize{9.8pt}{9pt}\selectfont
{\faGlobe ~}{\href{https://haroldchen19.github.io/PhysOPSD-site}{https://haroldchen19.github.io/PhysOPSD-site}}~~~
{\faGithub ~}{\href{https://github.com/aigc3d/PhysOPSD}{https://github.com/aigc3d/PhysOPSD}}
}

\begin{document}
\twocolumn[{
\renewcommand\twocolumn[1][]{#1}
\maketitle
\begin{center}
    \vspace{-22pt}
   \includegraphics[width=0.94\linewidth]{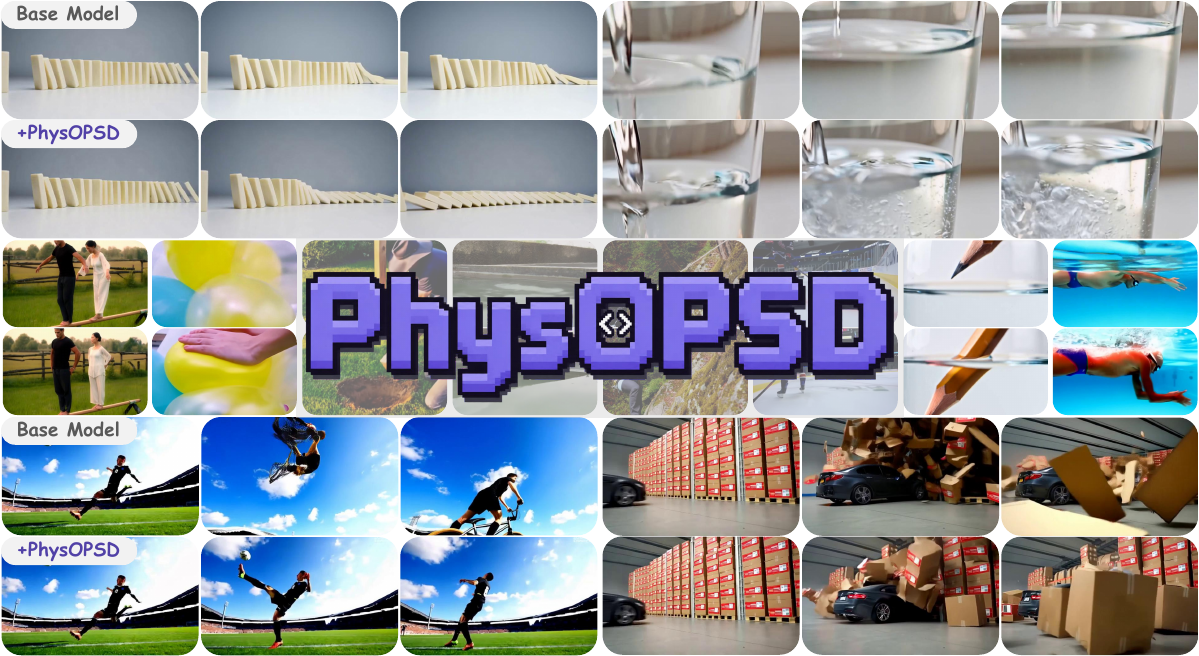}
   \vspace{-0.8em}
    \captionof{figure}{\label{fig:teaser}
    By internalizing structured physical privileges, \ourmethod empowers base models to generate more realistic motion trajectories and causal dynamic interactions across diverse physical regimes.}
\end{center}
}]

\begin{abstract}
While modern video generative models achieve remarkable visual fidelity, they remain noticeably brittle when generating physically plausible dynamics and complex interactions. 
Targeted post-training is crucial, yet supervised fine-tuning suffers from offline state mismatch, and reinforcement learning is hindered by sparse, ill-defined physical reward functions. 
On-Policy (Self-)Distillation (OPSD) reconciles on-policy rollouts with dense corrective feedback, shifting the core challenge to designing effective privileged context. 
However, conventional modalities fall short: text prompts are inherently declarative, whereas real reference videos heavily entangle physical laws with visual appearance without parametric control. 
To this end, we propose \ourmethod, introducing \textbf{executable code as privileged physical context} for video generation. 
Unlike static visual observations, executable code instantiates an editable physical world with intrinsic \textbf{abstraction} and \textbf{controllability}. 
Powered by \ourengine, we translate real-world physical scenarios into verifiable simulation programs. 
\ourmethod then incorporates (\textbf{\textit{i}}) dual-level physical privilege to bridge the information gap via an intermediate bottleneck, alongside (\textbf{\textit{ii}}) controlled physical variations under shared rollout noise to expose causal dynamic shifts. 
Extensive experiments demonstrate that \ourmethod substantially improves physical plausibility across diverse backbones, outperforming SFT and RL baselines with only $3$K samples, while preserving general video generation quality.
\end{abstract}

\vspace{-1.4em}
\section{Introduction}
\label{sec:intro}

Recent video generation models \citep{wan2025wan, hacohen2024ltx, kong2024hunyuanvideo, agarwal2025cosmos} have made impressive strides in visual fidelity and temporal coherence, yet they remain noticeably brittle when generating physically plausible dynamics and complex interactions \citep{motamed2026generative, bansal2025videophy, bansal2026videophy}. Targeted post-training has thus emerged as a vital route to reinforce physical capabilities beyond pre-training \citep{li2026video, puspitasari2024sora}. However, existing paradigms involve clear trade-offs: \ding{192} \textit{supervised fine-tuning (SFT)} offers dense supervision \citep{wang2026wisa, nan2025openvid, le2025gravity, yuan2026newtongen}, but operates on offline data that mismatches the states visited during inference; \ding{193} \textit{reinforcement learning (RL)} optimizes directly over on-policy rollouts \citep{kazimi2026hiphy, feng2026newton, ji2025physmaster}, yet suffers from sparse, ill-defined rewards and credit-assignment bottlenecks.

\begin{figure}[!t]
\centering
\includegraphics[width=\linewidth]{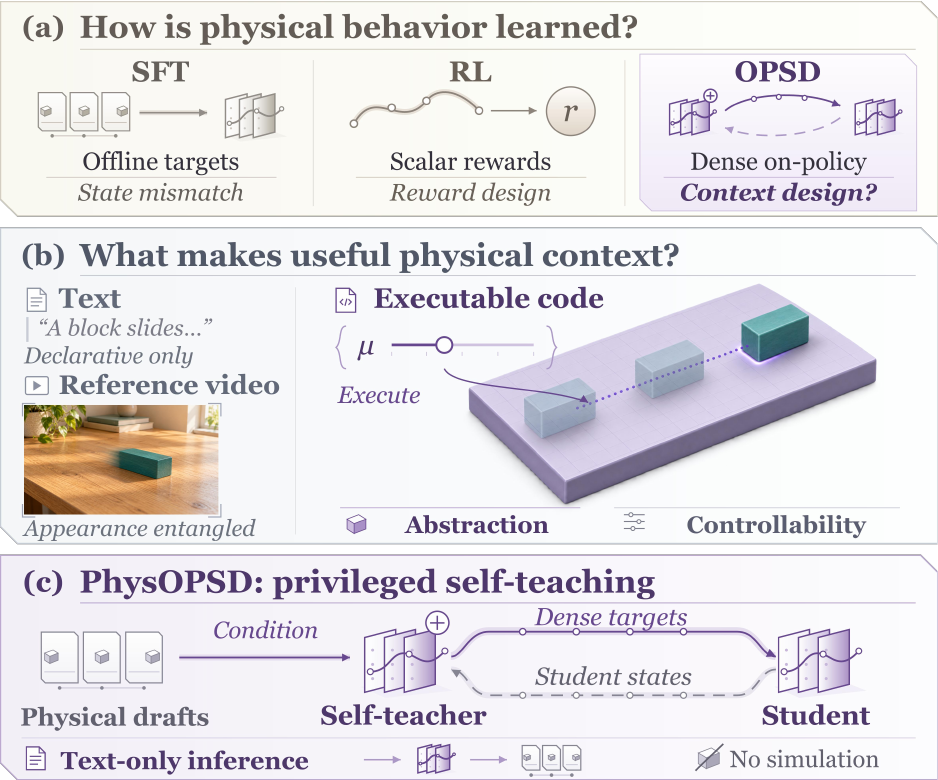}
\vspace{-1.8em}
\caption{\textbf{Background overview.} (\textbf{\textit{a}}) OPSD shifts physical post-training from reward engineering to privileged context design; (\textbf{\textit{b}}) Executable code explicitly parameterizes physical processes, uniquely unlocking \textit{abstraction} and \textit{controllability}; (\textbf{\textit{c}}) Using \textit{executable code as privileged context}, \ourmethod internalizes physical dynamics into the student for simulation-free video generation.}
\label{fig:figure1}
\vspace{-1.2em}
\end{figure}

Recent \ding{194} \textit{on-policy distillation (OPD)} \citep{li2026rethinking, song2026survey} provides a natural middle ground by querying a teacher for dense supervision on student-visited rollout states. Building upon this, \textbf{on-policy self-distillation (OPSD)} \citep{zhao2026self} further removes the need for an external teacher by casting the same model into asymmetric roles: the student acts on inference-time inputs, while the teacher accesses additional \textit{privileged context} to supervise the student rollouts. The central question of OPSD therefore becomes \textit{how to leverage extra information to turn the same model into a stronger self-teacher}, as shown in \Cref{fig:figure1} (\textit{a}). An effective privileged context should directly expose the information required by the target capability and reliably enhance teacher predictions on student rollout states. This principle has shown strong promise in language reasoning via ground-truth answers and rationales \citep{zhao2026self, penaloza2026privileged, yang2026self}, as well as in image generation via target-image multimodal representations \citep{jiang2026d}.

Extending this paradigm to video generation, however, is far less straightforward, as the primary bottleneck lies in modeling physically plausible dynamics, a capability that also defies evaluation by standard reward models \citep{li2026video, wang2026wisa, bansal2026videophy}. 
Yet, designing effective privileged context remains non-trivial: text expansion provides richer descriptions but remains fundamentally declarative, whereas real reference videos capture complete dynamics but heavily entangle them with irrelevant appearance, identity, and background clutter without controllable structure (\Cref{fig:figure1} (\textit{b})).
These bottlenecks lead to our core research question:
\vspace{-0.8em}
\begin{questionbox}[Research Question]
\textit{How can we construct privileged context that unlocks a video generator as its own physics teacher?}
\end{questionbox}
\vspace{-0.4em}
To answer this question, we introduce \ourmethod, which proposes \textit{\textbf{executable code as privileged physical context}} for video generation OPSD. Our key insight is that an executable physical scene instantiates an editable world rather than a fixed visual outcome.
This executability inherently provides two structural properties absent in raw video data: \ding{182}~\textbf{abstraction}, the ability to express dynamics across varying information densities; and \ding{183}~\textbf{controllability}, the ability to isolate and systematically perturb underlying physical parameters. To operationalize this insight, we first develop \ourengine, which converts diverse real-world scenarios into verifiable simulation code and renders synchronized physical drafts via a coding agent.

\ourmethod then systematically translates these two properties into OPSD:
\ding{168}~\textbf{dual-level physical privilege} exploits abstraction by pairing dense drafts with abstract drafts, establishing an intermediate physical information bottleneck to bridge the privilege gap and prevent appearance shortcuts;
\ding{171}~\textbf{controlled physical variation} exploits controllability by altering specific physical parameters across paired scenes with shared rollout noise, explicitly supervising how physical shifts dictate dynamic outcomes.
By internalizing structured code privilege through on-policy self-distillation, \ourmethod empowers base video generators to generate more physically plausible dynamics and stable interactions directly from text prompts.

To summarize, this work contributes threefold: 
\begin{itemize}[leftmargin=*]
    \item We propose code as privileged physical context for video generation, and introduce \ourmethod, an on-policy self-distillation framework incorporating \textit{dual-level physical privilege} to bridge the information gap and \textit{controlled physical variation} to capture causal dynamic shifts.
    \item We develop \ourengine to translate real-world physical scenarios into verifiable and controllable simulation programs, and construct \ourdata, an executable physical scene dataset featuring multi-abstraction drafts and paired physical variants.
    \item Extensive experiments demonstrate that \ourmethod consistently improves physical plausibility across diverse backbones, outperforming prior SFT and RL baselines with only a small fraction of their training data, without compromising general generation quality.
\end{itemize}

\section{Related Work}

\paragraph{On-Policy Distillation.}
On-policy distillation (OPD) mitigates train-test distribution shifts by supervising the student directly on its self-generated states with dense feedback \citep{li2026rethinking, song2026survey}. 
This paradigm has been applied to consolidate domain experts in language models such as MOPD \citep{ma2026mopd}, fuse visual capabilities in diffusion and flow models \citep{li2026diffusionopd, fang2026flow}, and optimize sampling trajectories in video generation \citep{gu2026anyflow}. 
To eliminate the reliance on external teachers, on-policy self-distillation (OPSD) constructs a stronger self-teacher from the same model under asymmetric privileged information \citep{zhao2026self}. 
While ground-truth solutions and rationales serve as privileged context in language reasoning \citep{zhao2026self, penaloza2026privileged, yang2026self}, D-OPSD \citep{jiang2026d} employs target-image multimodal representations to tune few-step image diffusion models. 
For video models, OPSD-V \citep{liu2026opsd} replaces older KV-cache history with real video frames to mitigate long-horizon drift in few-step autoregressive (AR) generators. 
Unlike OPSD-V, which focuses on causal AR cache stabilization, \ourmethod explores on-policy self-distillation for general video generation, introducing executable code as privileged physical context.

\vspace{-1em}
\paragraph{Code-for-Vision.}
Empowered by modern foundational vision-language models (VLMs) \citep{singh2025openai, team2023gemini}, executable code is increasingly utilized for visual scene synthesis and understanding. 
Early efforts such as 3D-GPT \citep{sun20253d} synthesize Blender scripts to procedurally build 3D environments \citep{kumaran2023scenecraft}. 
Recent studies further leverage coding agents for programmatic visual reconstruction, editing, and quantitative physical reasoning \citep{yin2026vision, li2026lego, tang2026bvb, wang2026code}. 
More recently, researchers have explored coding agents to assist visual generation by delegating spatiotemporal processes to code execution \citep{chen2025code2video}, either as physics-aware drafts in Blender \citep{li2026videococo} or cinematic previsualizations in Three.js \citep{liang2026codevideomodel}.
While existing generative approaches largely seek to replace part of generative modeling with coding agent capabilities, \ourmethod explores how executable code can empower the generative model itself.

\begin{figure*}[!t]
\centering
\includegraphics[width=\linewidth]{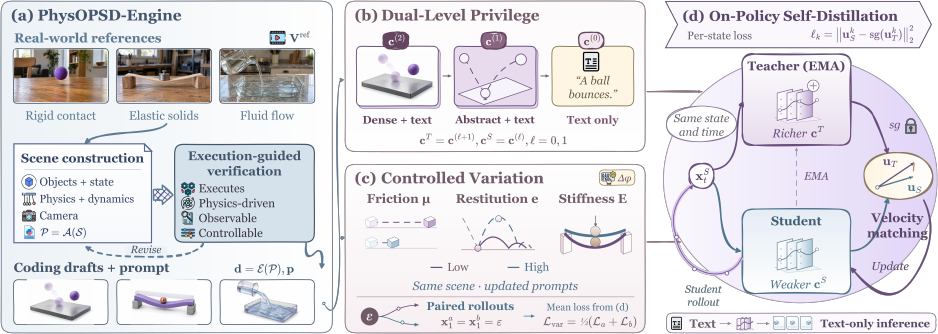}
\vspace{-1.8em}
\caption{Overview of our proposed \ourmethod.}
\label{fig:method}
\vspace{-0.6em}
\end{figure*}

\section{Preliminary: Problem Formulation}
\label{sec:preliminaries}

Modern video diffusion models \citep{wan2025wan, hacohen2024ltx, pan2026omniweaving} are predominantly formulated via flow matching \citep{lipman2022flow}, parameterizing a time-dependent velocity field:
\setlength\abovedisplayskip{3.4pt}
\setlength\belowdisplayskip{3.4pt}
\begin{equation}
\frac{\mathrm{d}\mathbf{x}_t}{\mathrm{d}t} = \mathbf{v}_{\bm{\theta}}(\mathbf{x}_t, t, \mathbf{p}),
\label{eq:flow_ode}
\end{equation}
where $\mathbf{x}_t$ denotes the video latent at continuous timestep $t \in [0, 1]$, $\mathbf{p}$ is the text prompt, and $\mathbf{v}_{\bm{\theta}}$ represents the neural velocity predictor with parameters $\bm{\theta}$. 
Starting from Gaussian noise $\mathbf{x}_1 \sim \mathcal{N}(\mathbf{0}, \mathbf{I})$, a video is generated over a discrete schedule $1 = t_K > t_{K-1} > \dots > t_0 = 0$ via a numerical ODE solver $\Phi$:
\begin{equation}
\mathbf{x}_{t_{k-1}} = \Phi(\mathbf{x}_{t_k}, t_k, t_{k-1}, \mathbf{u}^k), \quad \text{with } \mathbf{u}^k = \mathbf{v}_{\bm{\theta}}(\mathbf{x}_{t_k}, t_k, \mathbf{p}).
\label{eq:solver_step}
\end{equation}
This iterative integration yields the full generative trajectory that characterizes the model's rollout behavior.

On-policy self-distillation (OPSD) optimizes this generative process by querying a privileged self-teacher directly along the student's own rollout states. 
Formally, let $\mathbf{c}^S$ denote the student condition available at inference (\eg, prompt $\mathbf{p}$), and $\mathbf{c}^T$ the privileged teacher condition augmented with additional task-specific context. 
During training, the student first samples an on-policy rollout trajectory:
\begin{equation}
\bm{\tau}^S = \{\mathbf{x}_{t_K}^S, \mathbf{x}_{t_{K-1}}^S, \dots, \mathbf{x}_{t_0}^S\} \sim \mathcal{G}_{\bm{\theta}}(\mathbf{c}^S).
\label{eq:student_rollout}
\end{equation}
At each student-visited state $(\mathbf{x}_{t_k}^S, t_k)$, student and teacher predict their respective velocities:
\begin{equation}
\mathbf{u}_S^k = \mathbf{v}_{\bm{\theta}}(\mathbf{x}_{t_k}^S, t_k, \mathbf{c}^S), \qquad \mathbf{u}_T^k = \mathbf{v}_{\bar{\bm{\theta}}}(\mathbf{x}_{t_k}^S, t_k, \mathbf{c}^T),
\label{eq:velocity_predictions}
\end{equation}
where $\bar{\bm{\theta}}$ denotes the teacher parameters, instantiated as an EMA copy of $\bm{\theta}$. 
The student is then optimized by velocity matching over its self-generated rollout trajectory:
\begin{equation}
\mathcal{L}_{\mathrm{OPSD}} = \mathbb{E}_{\bm{\tau}^S \sim \mathcal{G}_{\bm{\theta}}(\mathbf{c}^S)} \left[ \frac{1}{K} \sum_{k=1}^K \left\| \mathbf{u}_S^k - \operatorname{sg}(\mathbf{u}_T^k) \right\|_2^2 \right],
\label{eq:opsd_loss}
\end{equation}
where $\operatorname{sg}(\cdot)$ is the stop-gradient operator. 
Crucially, the privileged context modifies only the teacher's conditioning, leaving the student's rollout states and inference behavior unaltered. 
The effectiveness of OPSD therefore hinges on whether the privileged context reliably improves teacher predictions on student-visited states.

\section{\ourmethod}
\label{sec:method}

\subsection{Overview}
\label{sec:overview}

OPSD relies on privileged context that reliably improves self-teacher predictions along student rollout states.
In modeling physical dynamics, however, conventional modalities fall short: text prompts remain inherently declarative, whereas real reference videos merely capture entangled visual observations.

\vspace{-1em}
\paragraph{Executable Code as Privileged Context.} 
\ourmethod bridges this gap by introducing executable code as privileged physical context, as shown in \Cref{fig:method}.
Unlike static text or video observations, an executable program explicitly specifies physical laws, object states, and dynamic parameters within an editable simulation, thereby uniquely providing explicit parametric control.
To operationalize this, \ourengine first translates real-world physical references $\mathbf{V}^{\mathrm{ref}}$ into an executable program $\mathcal{P}$. 
Simulating $\mathcal{P}$ in backend $\mathcal{E}$ yields a visual physical draft $\mathbf{d} = \mathcal{E}(\mathcal{P})$ and a grounded target prompt $\mathbf{p} = \operatorname{Describe}(\mathcal{P}, \mathbf{d})$. 
Here, code $\mathcal{P}$ generates the underlying physical dynamics, while the rendered visual draft $\mathbf{d}$ conditions the self-teacher.

\vspace{-1em}
\paragraph{Structured Physical Supervision.} 
\ourmethod builds upon an omni video generator $\mathbf{v}_{\bm{\theta}}(\mathbf{x}_t, t, \mathbf{p}, \mathbf{d})$, where $\mathbf{d} = \emptyset$ denotes text-only conditioned generation. 
Rather than treating $\mathbf{d}$ as a static condition, \ourmethod leverages two intrinsic properties of executable code: 
\begin{itemize}[leftmargin=1.4em,itemsep=1pt,topsep=1pt]
    \item[\ding{182}] \textbf{Abstraction}: synthesizing dense drafts $\mathbf{d}^{\mathrm{dense}}$ and abstract drafts $\mathbf{d}^{\mathrm{abs}}$ to establish a stepped hierarchy ($\mathbf{d}^{\mathrm{dense}} \to \mathbf{d}^{\mathrm{abs}} \to \mathbf{p}$) that smooths the privilege gap; 
    \item[\ding{183}] \textbf{Controllability}: perturbing physical parameters $\bm{\phi}$ across paired scenes with shared rollout noise to capture causal dynamic shifts. 
\end{itemize}
The subsequent \Cref{sec:engine} details \ourengine and \ourdata, while \Cref{sec:dual_privilege,sec:controlled_variation} elaborate the structured self-distillation mechanisms of \ourmethod.

\subsection{Executable Physical Context Engine}
\label{sec:engine}

To systematically synthesize verifiable and controllable privileged physical context, we develop \ourengine, an agentic pipeline that translates real-world physical references into executable scene programs and rendered physical drafts.

\vspace{-1em}
\paragraph{Reference-Guided Scene Construction.}
To seed simulations with natural diversity, \ourengine ingests real-world physical scenarios from WISA \citep{wang2026wisa} and public video collections. 
Rather than attempting to simulate arbitrary physical phenomena, the physical scope is deliberately restricted to \textit{simulation-groundable physical dynamics}, categorized into three domains: 
\textbf{(I)} rigid-body and contact dynamics, 
\textbf{(II)} elastic and deformable solid mechanics, and 
\textbf{(III)} fluid dynamics. 
Phenomena such as turbulent gas diffusion or thermodynamic phase shifts are excluded due to numerical instability and the difficulty of isolating causal physical parameters in standard simulators.

Given a reference video $\mathbf{V}^{\mathrm{ref}}$, a coding agent $\mathcal{A}$ then extracts the dominant physical event and formulates a structured physical scene specification:
\begin{equation}
\mathcal{S} = (\mathcal{O}, \mathcal{X}_0, \bm{\Phi}, \mathcal{E}, \mathcal{C}),
\label{eq:scene_spec}
\end{equation}
where $\mathcal{O}$ defines object geometries, $\mathcal{X}_0$ specifies initial kinetic states, $\bm{\Phi}$ collects physical properties (\eg, friction, elasticity, mass), $\mathcal{E}$ indicates interaction mechanisms, and $\mathcal{C}$ configures the camera trajectory. 
The coding agent then compiles $\mathcal{S}$ into an executable simulation program $\mathcal{P} = \mathcal{A}(\mathcal{S})$. 
Standard simulation backends such as MuJoCo \citep{todorov2012mujoco}, Genesis \citep{genesis2026genesisworld}, and Blender \citep{BlenderOnlineCommunity2026} are employed purely for their deterministic contact solvers and numerical stability, without asserting the superiority of any specific engine. 
Crucially, the governing principle is to \textit{reconstruct the physical scenario \textbf{rather than} visually reproduce the reference video}: geometries and textures are simplified as long as the underlying physical mechanics are faithfully instantiated.

\vspace{-1em}
\paragraph{Execution-Guided Verification.}
The synthesized program $\mathcal{P}$ is then refined through an iterative \textit{execution-observation-revision} loop. 
Rather than optimizing for pixel-level trajectory similarity to $\mathbf{V}^{\mathrm{ref}}$, verification enforces four strict criteria: 
\begin{itemize}[leftmargin=1em,itemsep=1pt,topsep=1pt]
    \item \textit{Execution validity}, ensuring code executes and renders without runtime divergence; 
    \item \textit{Physical validity}, verifying that dynamic events emerge from simulation mechanics rather than hard-coded keyframe animations; 
    \item \textit{Visual observability}, confirming that primary dynamic interactions are clearly visible within the viewport; 
    \item \textit{Parameter controllability}, ensuring that systematically perturbing key physical parameters produces physically consistent and observable dynamic responses;
\end{itemize}
Once verified, the target prompt is grounded directly on the executed outcome, $\mathbf{p} = \operatorname{Describe}(\mathcal{S}, \mathbf{d})$, guaranteeing semantic and causal alignment across $\mathbf{p} \leftrightarrow \mathcal{P} \leftrightarrow \mathbf{d}$.

\vspace{-1em}
\paragraph{\ourdata Dataset.} Using \ourengine, we further construct \textbf{\ourdata}, an executable physical scene dataset for video generation. 
Each sample is represented as a structured tuple:
\begin{equation}
\mathcal{D}_i = \{\mathbf{p}_i, \mathcal{P}_i, \mathbf{d}_i, \mathcal{M}_i\},
\label{eq:data_sample}
\end{equation}
where $\mathcal{P}_i$ is the executable program, $\mathbf{d}_i$ is the rendered visual draft, $\mathbf{p}_i$ is the grounded prompt, and $\mathcal{M}_i$ denotes structured metadata detailing physical parameters and execution configurations. 
Unlike static video collections, this executable formulation endows \ourdata with intrinsic \textit{abstraction} and \textit{controllability}. 
Encompassing the base scenes alongside their derived multi-abstraction drafts and parameter-controlled variants, \ourdata comprises $3$K total instances. 
We next detail how these structured properties are translated into self-distillation.

\begin{figure}[!t]
\centering
\includegraphics[width=\linewidth]{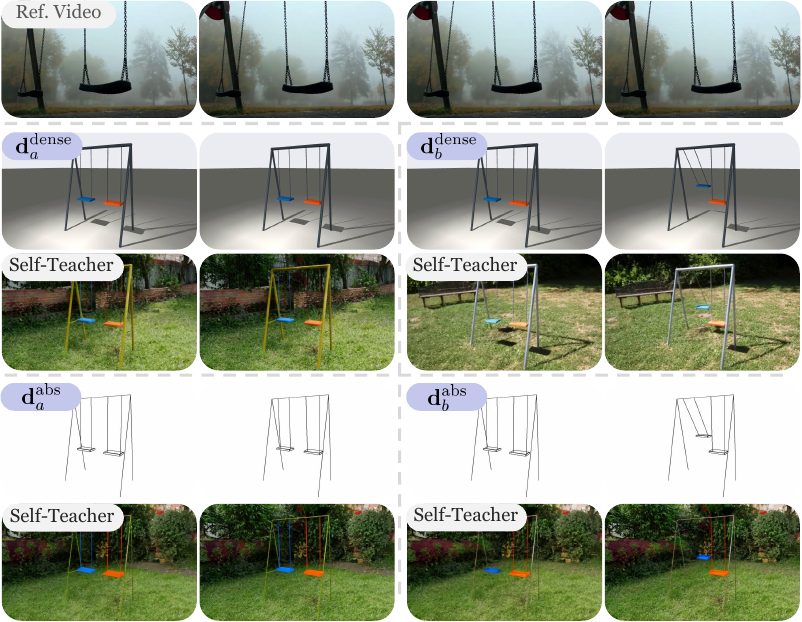}
\vspace{-1.8em}
\caption{\textbf{Case study on physical context and self-teacher generation.} 
From a swing-set reference, \ourmethod constructs dual-level drafts ($\mathbf{d}^{\mathrm{dense}}, \mathbf{d}^{\mathrm{abs}}$) across paired variants varying only in initial angular velocity ($0.12$ \vs $1.7\text{ rad/s}$). 
Conditioned on these drafts, the self-teacher faithfully synthesizes the corresponding motion dynamics. 
More case studies are in \Cref{app_fig:case_1,app_fig:case_2,app_fig:case_3}.}
\label{fig:case_study}
\vspace{-1em}
\end{figure}

\subsection{Structured Physical Self-Distillation}
\label{sec:structured_opsd}

Building upon the executable formulation, \ourmethod further translates code abstraction into \ding{168} \textit{dual-level physical privilege} to bridge the information gap, and leverages controllability via \ding{171} \textit{controlled physical variation} to supervise causal dynamic shifts, as illustrated in \Cref{fig:case_study}.

\subsubsection{Dual-Level Physical Privilege}
\label{sec:dual_privilege}

Directly distilling from a dense physical draft $\mathbf{d}^{\mathrm{dense}}$ to a text-only student may incur two fundamental obstacles: (\textbf{\textit{i}}) an excessively wide privilege gap that impedes knowledge internalization, and (\textbf{\textit{ii}}) the risk of geometric shortcuts where the teacher overfits to explicit geometry rather than abstracting causal dynamics. 
We thus introduce an abstract physical draft $\mathbf{d}^{\mathrm{abs}}$ as an intermediate \textbf{physical information bottleneck}, isolating motion and contact topology while filtering out fine-grained visual details.

\vspace{-1em}
\paragraph{Privilege Construction.}
From a single scene program $\mathcal{P}$, the simulation backend $\mathcal{E}$ renders physical drafts across distinct information densities:
\begin{itemize}[leftmargin=1.4em,itemsep=1pt,topsep=1pt]
    \item[\ding{192}] \textit{Dense Draft $\mathbf{d}^{\mathrm{dense}}$}: preserves complete surface geometry, collision contacts, dynamic interactions, and continuous spatiotemporal evolution.
    \item[\ding{193}] \textit{Abstract Draft $\mathbf{d}^{\mathrm{abs}}$}: retains coarse object-level motion, spatial trajectories, and contact events via simplified primitives, largely stripping appearance details.
\end{itemize}
These contexts define three hierarchical conditioning levels $\mathbf{c}^{(\ell)}$ for $\ell \in \{0, 1, 2\}$:
\begin{equation}
\mathbf{c}^{(0)} = \mathbf{p}, \quad \mathbf{c}^{(1)} = (\mathbf{p}, \mathbf{d}^{\mathrm{abs}}), \quad \mathbf{c}^{(2)} = (\mathbf{p}, \mathbf{d}^{\mathrm{dense}}),
\label{eq:conditioning_levels}
\end{equation}
where $\mathbf{c}^{(0)}$ corresponds strictly to the deployment-time text-only condition.

\begingroup
\setlength{\tabcolsep}{5.4pt}
\begin{table*}[!t]
\renewcommand{\arraystretch}{1.34}
  \centering
  \caption{Evaluation on physics-focused benchmarks, \ie, PhyGround \citep{lin2026phyground}, VideoPhy-2 \citep{bansal2026videophy}; and general quality benchmark, \ie, VBench \citep{huang2024vbench}. SA and PC denote Semantic Adherence and Physical Commonsense. We highlight the \textbf{best} and \underline{second best}.}
  \centering
  \vspace{-0.7em}
  \scriptsize
   \begin{tabular}{lcc|ccc ccccc ccc}
     \hlineB{2.5}
     \multirow{2}{*}{\textbf{Method}} & \multirow{2}{*}{\textbf{Size}} & \textbf{Post-Train} & \multicolumn{3}{c}{\textbf{PhyGround \citep{lin2026phyground}}} & \multicolumn{5}{c}{\textbf{VideoPhy-2 \citep{bansal2026videophy}}} & \multicolumn{3}{c}{\textbf{VBench \citep{huang2024vbench}}} \\
      \cline{4-14}
     &  & \textbf{Data} & \textbf{General} & \textbf{Physics} & \textbf{Overall} & \textbf{SA} & \textbf{PC} & \textbf{SA $\ge4$} & \textbf{PC $\ge4$} & \textbf{All} &  \textbf{Total} &  \textbf{Quality} &  \textbf{Semantic} \\
     \hlineB{1.5}
     Veo 3.1 \citep{wiedemer2025video} & - & & 71.9 & 66.5 & 69.2 & 4.39 & 4.06 & 84.4 & 80.5 & 68.9 & - & - & - \\
     Cosmos3-Nano \citep{agarwal2025cosmos} & 16B  &  & 65.5 & 64.9 & 65.2 & 3.86 & 4.49 & 66.3 & 90.2 & 60.4 & - & - & - \\
     HunyuanVideo-1.5 \citep{kong2024hunyuanvideo} & 8.3B  &  & 67.1 & 65.4 & 66.3 & 3.84 & 4.57 & 67.9 & 92.4 & 62.9 & 79.0 & 83.2 & 62.1 \\
     \hline
     UniVideo \citep{wei2026univideo} & 13B & & 65.7 & 64.0 & 64.9 & 3.85 & 4.57 & 68.2 & 92.6 & 63.1 & 83.5 & 84.4 & \textbf{80.0} \\
     $+$ PhyT2V \citep{xue2025phyt2v} & 13B & & 63.5 & \textbf{67.1} & 65.3 & \textbf{3.96} & 4.52 & 69.7 & 91.2 & 64.3 & 83.1 & 84.1 & 79.1 \\
     $+$ SFT \citep{wang2026wisa} & 13B & 80K & 65.1 & 64.5 & 64.8 & \textbf{3.96} & 4.52 & 69.5 & 91.2 & 64.0 & \underline{83.9} & \textbf{85.0} & \underline{79.7} \\
     $+$ PhysHPO \citep{chen2025hierarchical} & 13B & 21K & \underline{66.6} & 65.0 & \underline{65.8} & \underline{3.93} & \underline{4.60} & \underline{70.7} & \underline{93.2} & \underline{66.3} & 83.5 & \underline{84.6} & 79.1 \\
     \rowcolor{blue!10}
     \textbf{$+$ \ourmethod (Ours)} & 13B & 3K & \textbf{67.3} & \underline{65.6} & \textbf{66.5} & \textbf{3.96} & \textbf{4.61} & \textbf{71.7} & \textbf{93.6} & \textbf{67.3} & \textbf{84.0} & \textbf{85.0} & \underline{79.7} \\
     \hline
     OmniWeaving \citep{pan2026omniweaving} & 8.3B &  & 63.1 & 66.7 & 64.9 & 3.93 & 4.51 & 68.5 & 90.9 & 62.9 & 83.1 & \textbf{84.4} & 78.0 \\
     $+$ PhyT2V \citep{xue2025phyt2v} & 8.3B & & 66.9 & 65.2 & 66.1 & \textbf{3.97} & 4.53 & 69.9 & 91.2 & 64.5 & 83.0 & 84.0 & 78.9 \\
     $+$ SFT \citep{wang2026wisa} & 8.3B & 80K & 64.3 & \underline{67.9} & 66.1 & 3.85 & 4.57 & 68.4 & 92.6 & 63.3 & \underline{83.2} & 84.2 & 79.1 \\
     $+$ PhysHPO \citep{chen2025hierarchical} & 8.3B & 21K & \underline{67.3} & 66.7 & \underline{67.0} & 3.92 & \underline{4.60} & \underline{70.6} & \underline{93.2} & \underline{66.0} & \underline{83.2} & \underline{84.3} & \underline{79.2} \\
     \rowcolor{blue!10}
     \textbf{$+$ \ourmethod (Ours)} & 8.3B & 3K & \textbf{68.6} & \textbf{68.0} & \textbf{68.3} & \underline{3.95} & \textbf{4.61} & \textbf{71.4} & \textbf{93.4} & \textbf{67.0} & \textbf{83.3} & \underline{84.3} & \textbf{79.3} \\
     \hlineB{2.5}
   \end{tabular}
  \label{tab:main}
  \vspace{-0.8em}
\end{table*} 
\endgroup

\vspace{-1em}
\paragraph{Distillation Formulation.}
\ourmethod then establishes a stepped distillation hierarchy across adjacent information levels, $\mathbf{c}^{(2)} \to \mathbf{c}^{(1)} \to \mathbf{c}^{(0)}$:
\ding{192} \textit{Dense-to-Abstract} ($\ell = 1$), where the student rolls out on-policy under $\mathbf{c}^{(1)}$ and the teacher supervises under $\mathbf{c}^{(2)}$ to guide motion synthesis under reduced structural cues;
\ding{193} \textit{Abstract-to-Text} ($\ell = 0$), where the student rolls out under inference-time text $\mathbf{c}^{(0)}$ and the teacher provides corrective supervision under $\mathbf{c}^{(1)}$, serving as the primary pathway for directly boosting text-to-video physical fidelity.
Both transitions are optimized via a unified dual-level objective:
\begin{equation}
\mathcal{L}_{\mathrm{dual}} = \mathbb{E}_{\ell \sim P(\ell)} \left[ \mathcal{L}_{\mathrm{OPSD}}(\mathbf{c}^{(\ell)}, \mathbf{c}^{(\ell+1)}) \right], \quad \ell \in \{0, 1\},
\label{eq:dual_loss}
\end{equation}
where $\mathcal{L}_{\mathrm{OPSD}}(\mathbf{c}^S, \mathbf{c}^T)$ denotes the velocity-matching loss in Eq.~\eqref{eq:opsd_loss}, evaluated over student rollout $\bm{\tau}^S \sim \mathcal{G}_{\bm{\theta}}(\mathbf{c}^{(\ell)})$. 
This design enables the generator to progressively internalize spatiotemporal mechanics from dense simulations down to pure text inputs.

\subsubsection{Controlled Physical Variation}
\label{sec:controlled_variation}

While dual-level privilege bridges information densities, generators may still overfit to canonical motions rather than responding to physical parameter shifts.
A key advantage of executable code is its capability to surgically perturb underlying parameters while holding other scene factors strictly invariant. 
\ourmethod exploits this controllability to establish a \textbf{causal physical contrast}, explicitly exposing how physical shifts dictate dynamic consequences.

\vspace{-1em}
\paragraph{Controlled Variation Construction.}
Given an executable scene parameterized by physical properties, denoted as $W(\bm{\phi})$, perturbing specific physical parameters $\bm{\phi}$ constructs paired variants:
\begin{equation}
W_a = W(\bm{\phi}_a), \qquad W_b = W(\bm{\phi}_b).
\label{eq:paired_variants}
\end{equation}
Typical perturbations adjust friction coefficients, restitution, object mass, initial velocities, or stiffness, while strictly preserving object geometries, camera trajectories, lighting, and initial compositions. 
Executing both scenes yields corresponding draft-prompt pairs:
\begin{equation}
(\mathbf{p}_a, \mathbf{d}_a), \qquad (\mathbf{p}_b, \mathbf{d}_b).
\label{eq:paired_data}
\end{equation}
Crucially, the grounded prompts $\mathbf{p}_a$ and $\mathbf{p}_b$ explicitly describe the contrasting physical conditions, ensuring that the text-only student possesses sufficient semantic context to distinguish differing physical behaviors during deployment.

\vspace{-1em}
\paragraph{Paired On-Policy Rollout.}
To decouple the causal impact of physical variation from generative stochasticity, both variants share an identical initial Gaussian noise:
\begin{equation}
\mathbf{x}_1^a = \mathbf{x}_1^b \sim \mathcal{N}(\mathbf{0}, \mathbf{I}).
\label{eq:shared_noise}
\end{equation}
The student executes independent on-policy rollouts under their respective conditioning:
\begin{equation}
\bm{\tau}_a^S \sim \mathcal{G}_{\bm{\theta}}(\mathbf{c}_a^S; \mathbf{x}_1^a), \qquad \bm{\tau}_b^S \sim \mathcal{G}_{\bm{\theta}}(\mathbf{c}_b^S; \mathbf{x}_1^b),
\label{eq:paired_rollout}
\end{equation}
and is optimized over each rollout trajectory via:
\begin{equation}
\mathcal{L}_{\mathrm{var}} = \frac{1}{2} \left[ \mathcal{L}_{\mathrm{OPSD}}(\mathbf{c}_a^S, \mathbf{c}_a^T) + \mathcal{L}_{\mathrm{OPSD}}(\mathbf{c}_b^S, \mathbf{c}_b^T) \right],
\label{eq:var_loss}
\end{equation}
where each branch naturally integrates the dual-level privilege hierarchy from Eq.~\eqref{eq:dual_loss}. 
Shared initial noise purely suppresses task-irrelevant rollout variance, making paired responses more directly comparable.

In summary, \ourmethod leverages the intrinsic abstraction and controllability of executable code to internalize spatiotemporal physics and dynamic causality into the generator weights. 
At inference, all simulation components and physical drafts are discarded, leaving a general T2V or TI2V generator $\mathbf{v}_{\bm{\theta}}(\mathbf{x}_t, t, \mathbf{p}, \emptyset)$ with zero runtime overhead.

\section{Experiments}

In this section, we conduct experiments to answer the following research questions:
(\textbf{RQ1}) Does \ourmethod improve physically plausible video generation?
(\textbf{RQ2}) Does executable code provide effective privileged supervision?
(\textbf{RQ3}) How do structured privileges improve learning?

\begin{figure*}[!t]
\centering
\includegraphics[width=\linewidth]{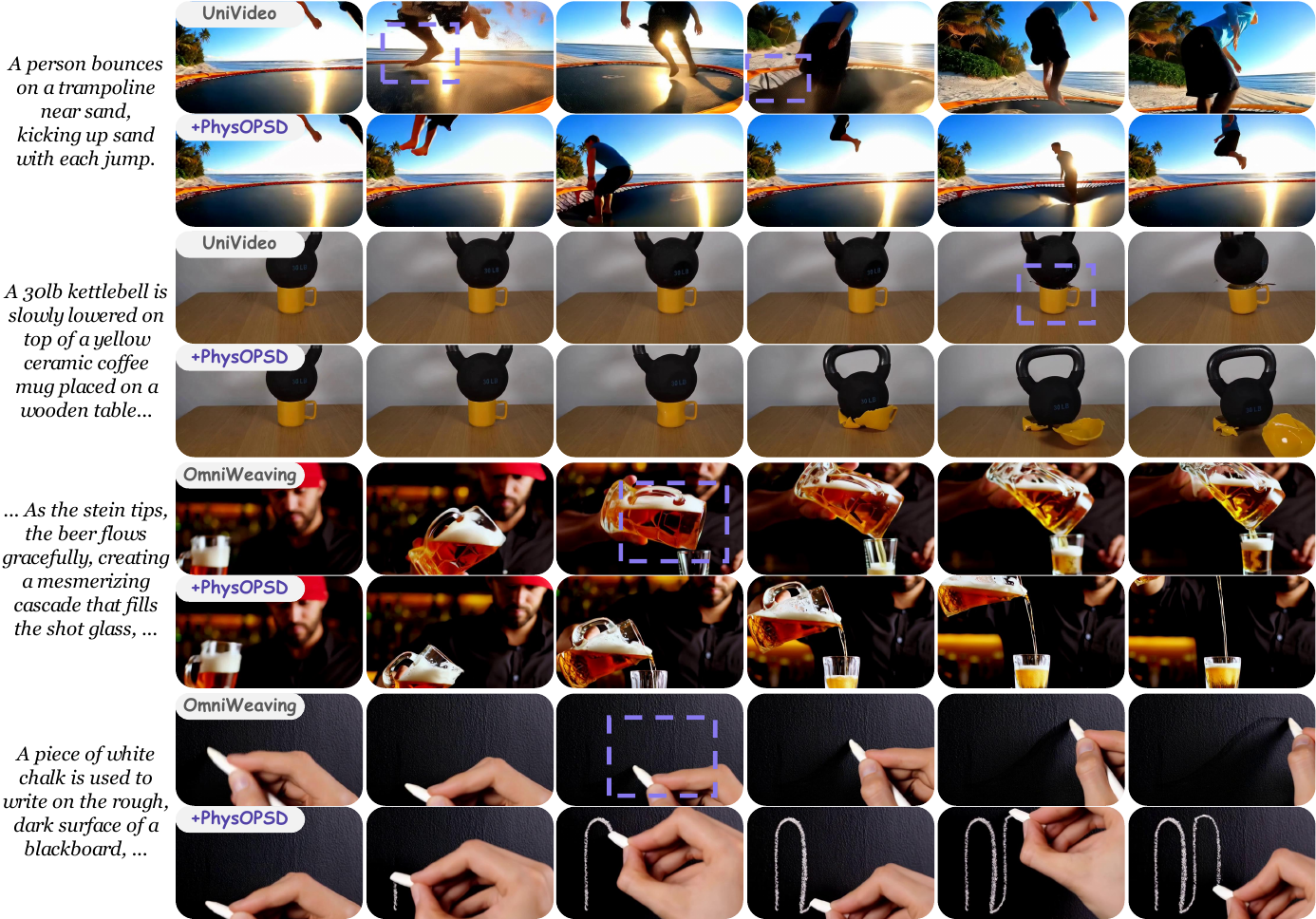}
\vspace{-1.8em}
\caption{Qualitative results of \ourmethod. More results are shown in \Cref{app_fig:qualitative_baseline_1,app_fig:qualitative_baseline_2,app_fig:qualitative_1,app_fig:qualitative_2}.}
\label{fig:qualitative}
\vspace{-0.6em}
\end{figure*}

\subsection{Experimental Settings}

\paragraph{Baselines.} We apply \ourmethod to two unified video generators, OmniWeaving \citep{pan2026omniweaving} and UniVideo \citep{wei2026univideo}. As a post-training method, \textit{our primary comparisons are against the respective base models} to directly measure capability gains. Since prior physics-oriented post-training methods are built on disparate backbones, we additionally implement three representative baselines: PhyT2V \citep{xue2025phyt2v} for test-time scaling, SFT using WISA \citep{wang2026wisa}, and PhysHPO \citep{chen2025hierarchical} for offline RL. We also report results from advanced video generators for reference, while further examining OPSD variants in our ablation studies. Details of baselines are in \Cref{app_sec:B}.

\vspace{-1em}
\paragraph{Evaluations.} We evaluate \ourmethod across two complementary aspects: 
\ding{192} \textit{physical plausibility}, adopting VideoPhy-2 \citep{bansal2026videophy} for T2V and PhyGround \citep{lin2026phyground} for TI2V to assess law-specific physical adherence; and 
\ding{193} \textit{general video quality}, employing VBench \citep{huang2024vbench} to measure general visual fidelity and semantic alignment.

\vspace{-1em}
\paragraph{Implementation Details.} 
\ourmethod optimizes student LoRA parameters using AdamW with a constant learning rate of $1\times 10^{-5}$, weight decay of $0.01$, and gradient clipping at $1.0$. 
Teacher weights are updated via an exponential moving average with a decay of $0.9999$ after paired gradient aggregation. 
Each student trajectory uses $K=20$ rollout denoising steps. 
Training employs BF16 mixed precision on NVIDIA H20 GPUs. 
At deployment, the student generates videos without teacher or simulation guidance.

\subsection{Main Results}

To answer \textbf{RQ1}, we conduct comprehensive evaluations of \ourmethod on physical and general generation benchmarks in \Cref{tab:main}, complemented by qualitative baseline comparisons (\Cref{app_fig:qualitative_baseline_1,app_fig:qualitative_baseline_2}) and diverse generative showcases (\Cref{fig:qualitative,app_fig:qualitative_1,app_fig:qualitative_2}).
Observations are summarized below:

\vspace{0.4em}
\noindent\textbf{Obs.\ding{182} \ourmethod consistently improves physical plausibility across backbones with high sample efficiency, while preserving general video quality.} 
As shown in \Cref{tab:main}, \ourmethod achieves top physical adherence across both UniVideo (13B) and OmniWeaving (8.3B), boosting VideoPhy-2 overall scores by $+4.2$ and $+4.1$, and PhyGround by $+1.6$ and $+3.4$, respectively, with physical commonsense (PC) reaching $4.61$. 
Crucially, \ourmethod achieves these gains using only $3$K samples, markedly outperforming SFT ($80$K) and offline RL ($21$K). 
Meanwhile, this physical specialization incurs no catastrophic forgetting: VBench confirms that \ourmethod maintains or even slightly improves overall visual quality and text alignment (\eg, UniVideo rises from $83.5$ to $84.0$, while OmniWeaving remains steady at $83.3$), verifying that on-policy distillation protects foundational visual representations.

\vspace{0.4em}
\noindent\textbf{Obs.\ding{183} \ourmethod generates physically faithful dynamics over baselines and generalizes robustly to visually rich, open-domain scenes.} 
In \Cref{app_fig:qualitative_baseline_1,app_fig:qualitative_baseline_2}, baselines exhibit motion stagnation, severe geometric distortions, or broken reflection mechanics during concrete leveling and teapot rotation, whereas \ourmethod synthesizes continuous, physically consistent interactions. 
Furthermore, although physical drafts in \ourengine use simplified simulation geometry, the learned dynamics transfer robustly to open-domain photorealistic scenes (\Cref{fig:qualitative,app_fig:qualitative_1,app_fig:qualitative_2}), successfully capturing fluid sloshing, impact cratering, elastic penetration, and optical reflection. 
This verifies that \ourmethod internalizes transferable physical mechanics rather than overfitting to synthetic simulation appearances.

\subsection{Analysis of Code-as-Privilege}
\label{sec:analysis_privilege}

To answer \textbf{RQ2}, we compare different privileged modalities under an identical budget of $3$K samples, as shown in \Cref{fig:analysi_privilege}: 
(\textbf{\textit{i}}) vanilla model (w/o Privilege); 
(\textit{\textbf{ii}}) declarative prompt expansion (w/ Refined Text); 
(\textbf{\textit{iii}}) real reference videos (w/ Real Video); and 
(\textbf{\textit{iv}}) executable simulation drafts (w/ Coding Draft). 
We give the following insights:

\begin{figure}[!h]
\centering
\includegraphics[width=\linewidth]{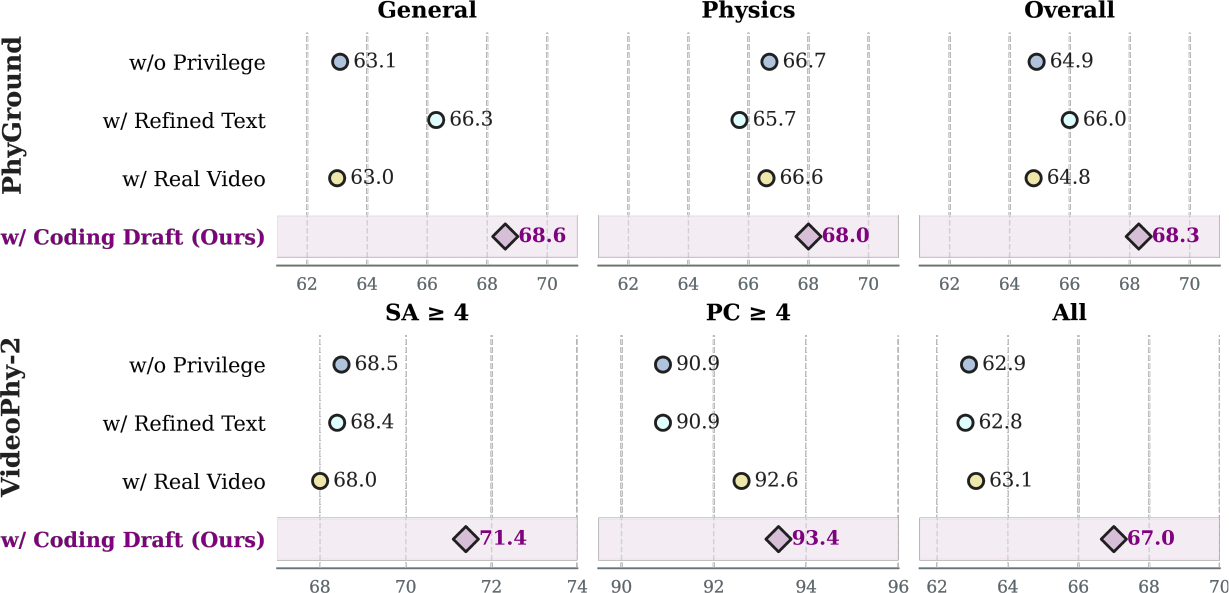}
\vspace{-2em}
\caption{\textbf{Privilege context analysis.} Executable coding drafts yield superior physical adherence over text and real videos.}
\label{fig:analysi_privilege}
\vspace{-1.2em}
\end{figure}

\vspace{0.4em}
\noindent\textbf{Obs.\ding{184} Executable code serves as an effective privileged context for physically plausible video generation.} 
As shown in \Cref{fig:analysi_privilege}, refined text (w/ Refined Text) yields negligible physical gains and slightly decreases VideoPhy-2 All ($62.8$ \vs $62.9$), showing that declarative descriptions cannot ground spatiotemporal mechanics. 
Meanwhile, real reference videos (w/ Real Video) produce only marginal improvements ($63.1$ \vs $62.9$), as complex background and visual clutter encourage the teacher to rely on visual shortcuts rather than transferable dynamics. 
In contrast, \ourmethod (w/ Coding Draft) achieves decisive improvements, boosting PhyGround Overall to $68.3$ ($+3.5$ over real video) and VideoPhy-2 All to $67.0$ ($+3.9$ over real video). 
As qualitatively shown in \Cref{fig:case_study,app_fig:case_1,app_fig:case_2,app_fig:case_3}, \ourmethod isolates pure dynamics from visual noise and enables parametric control, empowering the self-teacher with dense, physically faithful supervision.

\subsection{Analysis of Structured Privilege}
\label{sec:analysis_structure}

To answer \textbf{RQ3}, we further dissect the individual contributions of our two structural strategies under a controlled budget of $3$K samples: \textit{dual-level physical privilege} and \textit{controlled physical variation}. 
Performance gains over the base model on VideoPhy-2 are illustrated in \Cref{fig:analysi_structure}. We summarize the following observations:

\begin{figure}[!h]
\centering
\vspace{-0.6em}
\includegraphics[width=\linewidth]{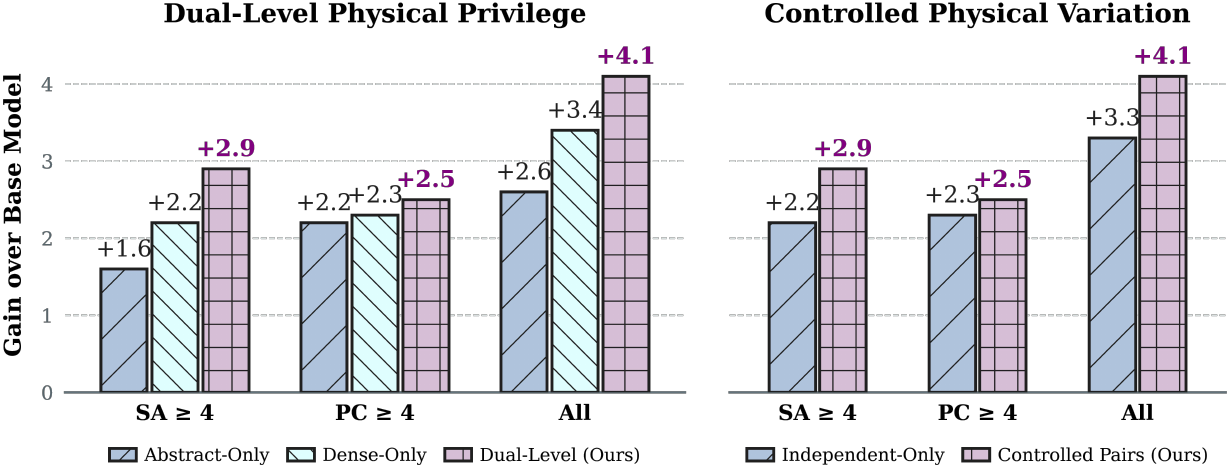}
\vspace{-2em}
\caption{\textbf{Structured physical privilege analysis.} Both (\textbf{\textit{Left}}) the stepped dual-level hierarchy and (\textbf{\textit{Right}}) parameter-controlled paired variations consistently yield superior physical adherence over their uncoupled counterparts.}
\label{fig:analysi_structure}
\vspace{-0.6em}
\end{figure}

\noindent\textbf{Obs.\ding{185} Dual-level privilege smooths knowledge transfer via an intermediate information bottleneck.} 
In \Cref{fig:analysi_structure} (\textit{Left}), while Abstract-Only ($+2.6$ on All) lacks complete geometric constraints and Dense-Only ($+3.4$) encourages visual shortcuts, our stepped hierarchy (Dual-Level) attains the highest gain ($+4.1$ on All). 
This confirms that abstract drafts effectively bridge the privilege gap to enable progressive dynamic internalization.

\vspace{0.4em}
\noindent\textbf{Obs.\ding{186} Controlled physical variation exposes dynamic causality beyond learning from isolated scenes.} 
In \Cref{fig:analysi_structure} (\textit{Right}), training on independent physical scenes without parameter-controlled variations (Independent-Only) achieves a $+3.3$ gain on All. 
In contrast, structuring data into paired variants differing by a single physical parameter (Controlled Pairs) elevates the gain to $+4.1$. 
This demonstrates that exposing explicit counterfactual dynamic shifts is far more effective for internalizing physical causality than merely observing standalone physical scenarios.

\subsection{Analysis of Data Efficiency}
\label{sec:analysis_data}

Finally, we examine the data efficiency and scaling behavior of \ourmethod by varying the training data budget from $1$K to $4$K instances. 
Performance gains are illustrated in \Cref{fig:analysi_data}:

\begin{figure}[!h]
\centering
\vspace{-0.6em}
\includegraphics[width=\linewidth]{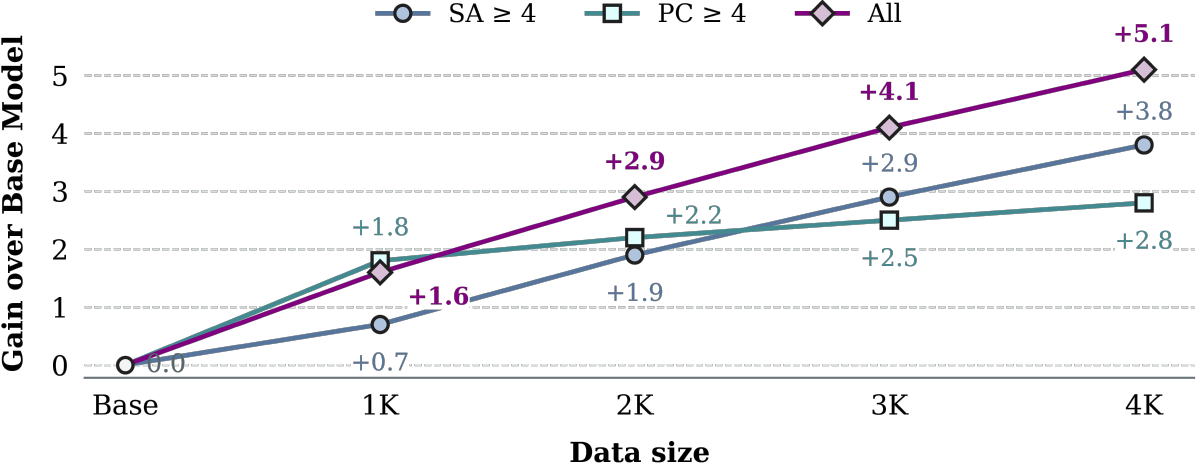}
\vspace{-2em}
\caption{\textbf{Data scaling analysis.} \ourmethod delivers physical improvements even in low-data regimes and scales steadily.}
\label{fig:analysi_data}
\vspace{-1em}
\end{figure}

\noindent\textbf{Obs.\ding{187} \ourmethod exhibits remarkable sample efficiency and steady scaling potential.} 
As shown in \Cref{fig:analysi_data}, with merely $1$K samples, \ourmethod yields an immediate improvement of $+1.6$ on All ($+1.8$ on $\text{PC} \ge 4$). 
As the dataset scales to $4$K, overall gains steadily increase to $+5.1$ ($+3.8$ on $\text{SA} \ge 4$ and $+2.8$ on $\text{PC} \ge 4$) without premature plateauing. 
Compared to traditional paradigms requiring tens of thousands of samples (\eg, $80$K for SFT and $21$K for offline RL), \ourmethod leverages dense on-policy supervision from structured code privileges, delivering high sample efficiency and sustainable scaling performance.

\section{Conclusion}

In this work, we present \ourmethod, a post-training framework that introduces executable code as privileged physical context for on-policy self-distillation. 
Recognizing that executable code instantiates an editable physical world with intrinsic abstraction and controllability, we develop \ourengine to construct verifiable simulation scenes and rendered drafts. 
\ourmethod systematically translates these properties into dual-level physical privilege to bridge the information gap via an intermediate bottleneck, alongside controlled physical variations under shared rollout noise to expose dynamic causality. 
Extensive experiments on standard physics benchmarks demonstrate that \ourmethod substantially improves physical plausibility across diverse backbones with high sample efficiency, while preserving general video quality. 
We hope our code-as-privilege inspires future research on leveraging coding agents to empower generative models rather than replacing them, advancing physically grounded visual generation.

\section{Scope of This Report}

\paragraph{Algorithm over Monolithic Models.} 
\ourmethod contributes a general on-policy self-distillation framework rather than delivering a single weight-locked model. 
This motivates our focus on relative capability gains over respective base generators (\eg, UniVideo \citep{wei2026univideo} and OmniWeaving \citep{pan2026omniweaving}) to confirm cross-architecture generalizability. 
Furthermore, our compact default budget of $3$K samples highlights sample efficiency, while our scaling analysis (\Cref{sec:analysis_data}) confirms steady performance gains as data increases.

\vspace{-1em}
\paragraph{Empowering Rather Than Replacing Generation.} 
Central to our philosophy is leveraging coding agents to empower generative models rather than substituting them at test time. 
\ourengine is thus designed as an open, modular pipeline interfacing standard coding LLMs with physics engines. 
As foundational coding agents and physics solvers advance, the pipeline naturally inherits the ability to synthesize increasingly complex physical scenarios without altering the distillation formulation.

\vspace{-1em}
\paragraph{Boundaries of Simulation-Groundable Dynamics.} 
The physical scope of \ourmethod is bounded by scenarios that standard simulation engines can stably compute: rigid-body contacts, elastic mechanics, and fluid dynamics. 
Chaotic multi-physics and thermodynamic phase transitions (\eg, combustion) are deliberately excluded due to solver instability and difficulty in isolating parameters. 
Rather than emulating all physical phenomena, our objective is establishing a principled blueprint for internalizing physical dynamics via code privilege.


\clearpage
{
    \small
    \bibliographystyle{ieeenat_fullname}
    \bibliography{main}
}

\clearpage
\appendix
\setcounter{page}{1}

\section{More Details of \ourmethod}
\label{app_sec:A}

\subsection{\ourengine Details}
\label{app_sec:A_1}

Synthesizing executable physical scenes requires translating abstract dynamic descriptions into syntactically valid and numerically stable simulation code. 
While frontier LLMs/VLMs exhibit proficient code generation and spatial reasoning, the non-linear nature of contact dynamics makes single-pass zero-shot generation unreliable for complex physical phenomena. 
To guarantee a verifiable physical basis, \ourengine combines a frontier coding agent with an automated execution-guided refinement loop.

\vspace{-1em}
\paragraph{Iterative Scene Refinement Workflow.} 
We adopt GPT-6-Astra as the backbone coding agent. 
Upon generating an initial scene program $\mathcal{P}$, \ourengine executes it in the target backend (\eg, MuJoCo \citep{todorov2012mujoco} or Blender \citep{BlenderOnlineCommunity2026}) and aggregates diagnostic feedback across three modalities: \ding{182} execution logs (\eg, syntax errors or solver convergence warnings), \ding{183} kinematic telemetry (\eg, penetration anomalies or constraint forces), and \ding{184} rendered preview keyframes. 
When physical implausibility, solver divergence, or poor camera observability is detected, the coding agent inspects the diagnostic trace to revise geometric configurations, contact parameters, or camera trajectories. 
This execution-observation-revision cycle iterates until all four verification criteria in \Cref{sec:engine} are satisfied.

\vspace{-1em}
\paragraph{Multi-Level Draft Generation.} 
To prevent dynamic discrepancies across privilege levels, dense drafts $\mathbf{d}^{\mathrm{dense}}$ and abstract drafts $\mathbf{d}^{\mathrm{abs}}$ for a given variant are \textit{rendered from an identical, frozen simulation state trajectory}. 
Specifically, once a simulation execution succeeds, \ourengine records the full kinematic trajectory of all bodies and contact events. 
$\mathbf{d}^{\mathrm{dense}}$ is rendered with full surface meshes, material shaders, and realistic lighting, whereas $\mathbf{d}^{\mathrm{abs}}$ renders simplified wireframe primitives and centroid trajectories from the exact same camera viewpoint. 
This trajectory freezing guarantees that both privilege levels share identical underlying dynamics, completely eliminating confounding dynamic drift caused by re-simulation. 
For controlled variants, non-target attributes and initial conditions are frozen to isolate the causal parameter shift.

\begin{figure}[!t]
\centering
\includegraphics[width=\linewidth]{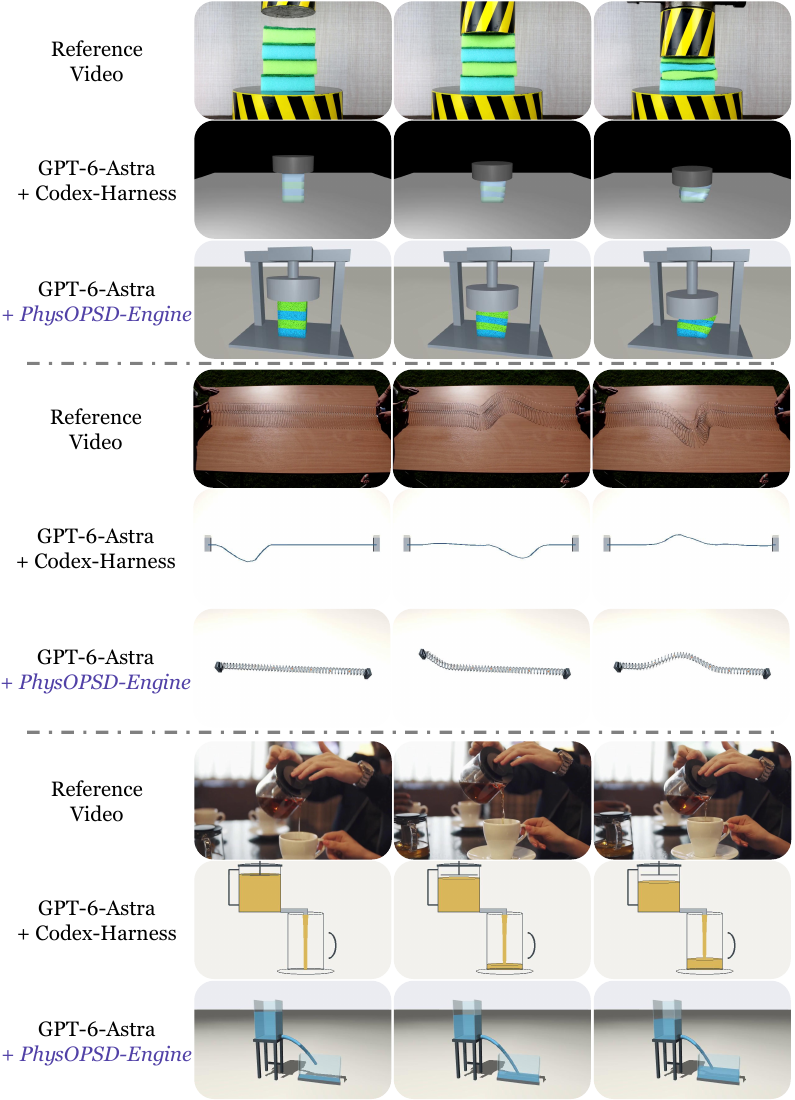}
\vspace{-2em}
\caption{Comparison of \ourengine \textit{vs.} Codex-Harness.}
\label{app_fig:harness}
\vspace{-1em}
\end{figure}

\vspace{-1em}
\paragraph{Outcome-Grounded Dual Prompt Annotation.} 
In the main text, we denote the target text condition uniformly as $\mathbf{p}$ for mathematical clarity. 
In practice, accommodating our omni-capable video generator entails producing two complementary text conditions via Qwen3-VL-32B \citep{yang2025qwen3}, conditioned on the dense draft alongside a structured fact card containing measured physical telemetry:
\begin{itemize}
    \item \textbf{T2V Prompt}: a self-contained description detailing the scene context, physical conditions, observable dynamic trajectory, and final state, used for deployment-time text-only student generation.
    \item \textbf{V2V Instruction}: an imperative prompt instructing the model to perform photorealistic visual refinement while strictly preserving the underlying spatiotemporal dynamics and spatial layout, shared by the teacher and draft-conditioned student.
\end{itemize}
Both text modalities are grounded strictly in the executed simulation telemetry rather than speculative prior descriptions. 
The concrete system prompts utilized for annotation are detailed below.

\begin{figure*}[t]
\centering
\begin{tcolorbox}[
    sharp corners,
    colframe=headingpurple!20,
    colback=gray!4,
    boxrule=3pt,
    boxsep=0.5pt,
    enhanced,
    width=\textwidth,
    shadow={3pt}{-3pt}{0pt}{opacity=1,gray!10},
    title={System Prompt for Dual Prompt Annotation}
]
\footnotesize
\setstretch{1}
{\fontfamily{pcr}\selectfont
\begin{lstlisting}[breaklines=true]
"""
You are an expert video annotation assistant specializing in physical dynamics, object interactions, and temporally grounded video descriptions. Your task is to generate accurate training prompts for text-to-video (T2V) generation and video-to-video (V2V) editing.

You are given two videos representing different physical settings of the same scene, together with chronological visual observations and a verified fact card. Treat them as independent physical outcomes, not as a before-and-after sequence. Prioritize factual accuracy, physical consistency, and temporal grounding over narrative richness.

1. VISUAL AND PHYSICAL GROUNDING

- Identify the principal objects, initial configuration, physical interactions, motion progression, and actual final states.
- Examine the full temporal sequence. Compare the first and last frames to distinguish temporary deformation from permanent changes, motion from detachment, and partial events from completed events.
- Describe observed dynamics rather than expected textbook outcomes. Changing a physical parameter does not guarantee a specific result.
- Use the verified fact card to identify physical conditions, affected objects, measured states, and camera settings. Ground the actual motion and outcomes in the provided visual evidence.
- Do not infer screen-space directions from numerical coordinates. Motion alone does not establish collision, fracture, or detachment.
- Distinguish physical quantities accurately, including friction, damping, mass, stiffness, and velocity. Do not confuse physical speed with video playback speed.
- Omit uncertain details rather than inventing events or outcomes.

2. ANNOTATION REQUIREMENTS

Generate two English training descriptions for EACH physical variant.

A. T2V Prompt (t2v_prompt): Write a self-contained target-video description, normally 70-115 words, including:
- Principal objects, geometry, spatial layout, and realistic appearance.
- Relevant qualitative physical conditions and affected objects.
- Actual motion, key interactions, and observed final outcomes.
- Camera viewpoint and scene composition.

The description must stand alone without referring to the source video, simulation, editing process, or the other variant.

B. V2V Instruction (v2v_instruction): Write a concise appearance-editing instruction, normally 35-65 words, using two sentences:
- Begin with "Convert this video into photorealistic live-action footage of ..." to specify realistic objects, materials, and appearance.
- Begin the second sentence with "Preserve ..." to maintain the physical conditions, motion dynamics, event sequence, final outcome, timing, spatial arrangement, and camera behavior.

The SAME V2V instruction must be used for both dense and abstract video representations of the same physical variant.

3. CONSISTENCY CONSTRAINTS

- Keep target materials and non-intervened scene attributes consistent across variants and between T2V and V2V descriptions.
- Preserve actual objects, apparatus, supports, containers, openings, and spatial relationships. Distinguish transparent containers from their liquid contents where applicable.
- Do not introduce unobserved objects, people, backgrounds, equipment, or physical events. Do not copy arbitrary proxy colors or textures.
- Respect the verified camera configuration. Changes in framing do not necessarily indicate camera motion.
- Do not include raw numerical coefficients, measurement units, variable names, pixel dimensions, or simulator-specific terminology in the training descriptions. Physical object counts are allowed.
- Describe each variant independently. Avoid relative expressions such as "faster than the other video"; describe its actual behavior instead.
- Ground descriptions in the CURRENT executed outcome, not in an original reference caption or an assumed physical result.

4. OUTPUT FORMAT ... 5. FINAL VERIFICATION ...

"""
\end{lstlisting}
}
\end{tcolorbox}
\end{figure*}

\vspace{-1em}
\paragraph{Effectiveness of \ourengine}

To further validate the necessity of our agentic execution-guided loop, we compare \ourengine against a standard harness framework (\ie, Codex) using the identical backbone coding model (GPT-6-Astra). 
As illustrated in \Cref{app_fig:harness}, naive code generation without closed-loop simulation feedback frequently degenerates into non-physical shortcuts: 
collapsing into 2D schematic toy drawings without fluid mechanics (\eg, fluid pouring), omitting structural apparatus and floating in an ungrounded void (\eg, sponge compression), or approximating continuous 3D dynamics with simplistic line segments (\eg, transverse spring waves). 
In contrast, \ourengine leverages multi-modal execution diagnostics and kinematic telemetry to iteratively refine boundary conditions, contact constraints, and spatial geometry. 
This closed-loop refinement guarantees that the synthesized programs instantiate verifiable 3D physical environments with observable, physically grounded dynamics.

\subsection{\ourdata Statistics and Case Studies}
\label{app_sec:A_2}

\begin{figure}[!h]
\centering
\vspace{-0.6em}
\includegraphics[width=\linewidth]{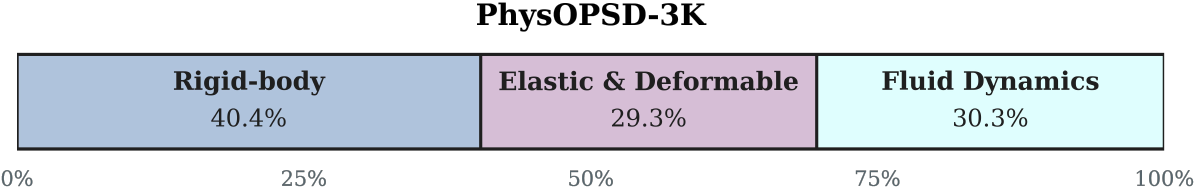}
\vspace{-2em}
\caption{Distribution of \ourdata.}
\label{app_fig:data}
\vspace{-1em}
\end{figure}

\paragraph{Dataset Statistics.} 
\ourdata comprises $3$K total instances, encompassing base executable scenes alongside their derived multi-abstraction drafts ($\mathbf{d}^{\mathrm{dense}}, \mathbf{d}^{\mathrm{abs}}$) and parameter-controlled paired variants ($\mathbf{d}_a, \mathbf{d}_b$). Distribution across domains are shown in \Cref{app_fig:data}.

\begin{figure*}[!t]
\centering
\includegraphics[width=\linewidth]{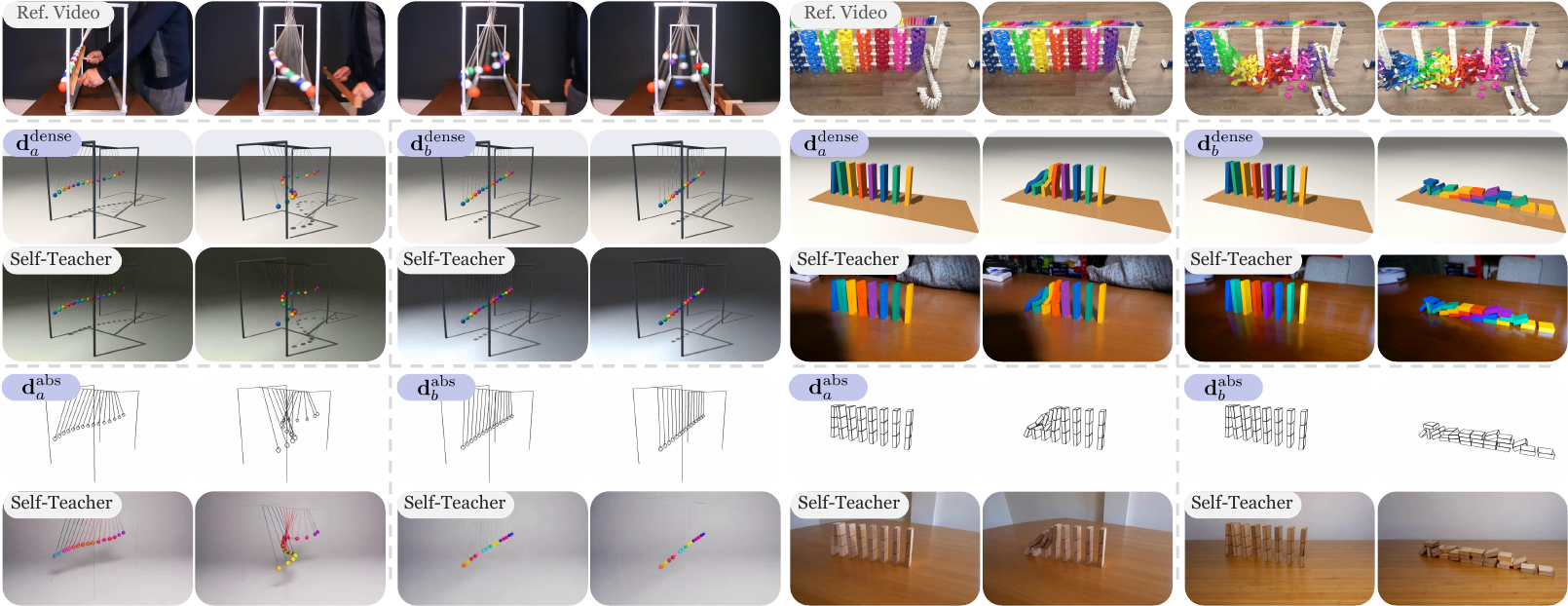}
\vspace{-2em}
\caption{Case study of \ourmethod on rigid-body and contact dynamics.
(\textbf{\textit{Left}}) \textit{Pendulum wave}: perturbing the damping rate ($0.018$ \vs $1.4\text{ s}^{-1}$) yields sustained wave oscillations \vs rapid kinetic decay. 
(\textbf{\textit{Right}}) \textit{Domino cascade}: varying gravitational acceleration ($0.981$ \vs $9.81\text{ m/s}^2$) alters the collapse speed from sluggish, incomplete toppling to rapid, sequential falling.}
\label{app_fig:case_1}
\vspace{-0.4em}
\end{figure*}

\begin{figure*}[!t]
\centering
\includegraphics[width=\linewidth]{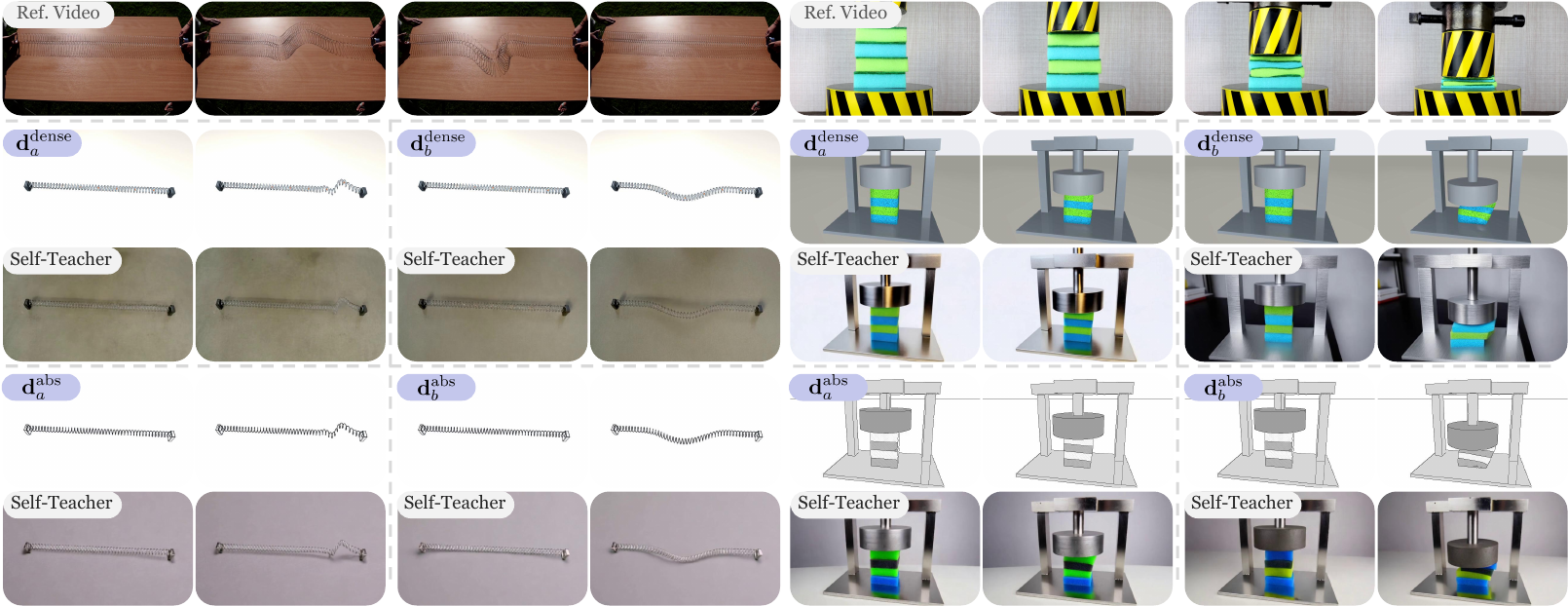}
\vspace{-2em}
\caption{Case study of \ourmethod on elastic and deformable solid mechanics. (\textbf{\textit{Left}}) \textit{Transverse spring wave}: modulating transverse coupling stiffness ($0.5$ \vs $8\text{ N/m}$) shifts dynamic response from localized dispersion to cohesive wave propagation. 
(\textbf{\textit{Right}}) \textit{Sponge compression}: increasing actuator displacement ($8$ \vs $24\text{ mm}$) drives the deformation from moderate compaction to severe, non-linear flattening.}
\label{app_fig:case_2}
\vspace{-0.4em}
\end{figure*}

\begin{figure*}[!t]
\centering
\includegraphics[width=\linewidth]{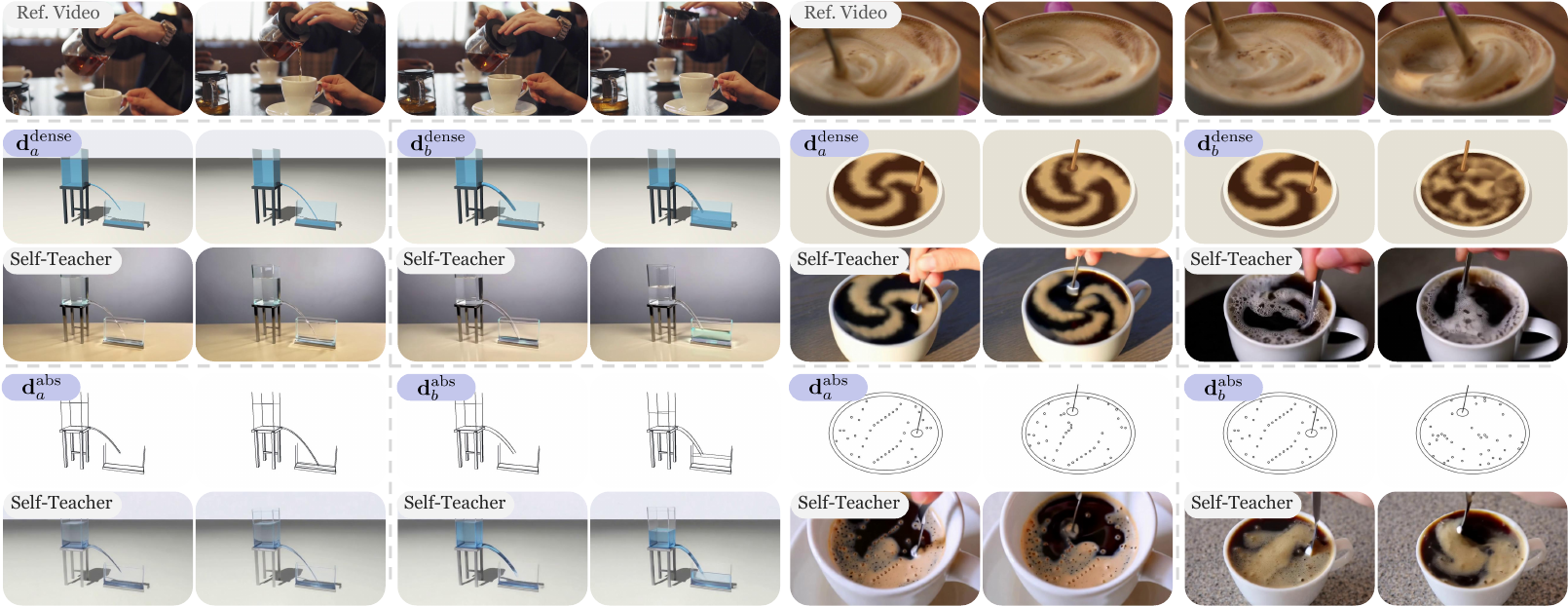}
\vspace{-2em}
\caption{Case study of \ourmethod on fluid dynamics. (\textbf{\textit{Left}}) \textit{Fluid drainage}: altering the container outlet radius ($1.8$ \vs $5.5\text{ mm}$) controls the discharge flux from a thin, laminar stream to a high-volume outflow. (\textbf{Right}) \textit{Coffee stirring}: varying spoon angular velocity ($3$ \vs $14\text{ rad/s}$) transitions the fluid motion from a smooth spiral vortex to vigorous turbulent mixing and foam dispersion.}
\label{app_fig:case_3}
\vspace{-0.4em}
\end{figure*}

\vspace{-1em}
\paragraph{Case Studies.} 
\Cref{app_fig:case_1,app_fig:case_2,app_fig:case_3} showcase qualitative case studies across each dynamic domain. 
By systematically perturbing underlying physical parameters (\eg, damping, gravity, stiffness, displacement, outlet radius, and stirring velocity), the self-teacher reliably synthesizes physically consistent dynamics under both dense geometric and abstract wireframe draft conditioning. 
These results confirm that code-derived privileges effectively isolate dynamic mechanics from visual noise, providing transferable physical guidance without appearance shortcuts.

\subsection{\ourmethod Algorithmic Overview}
\label{app_sec:A_3}

\ourmethod unifies executable physical context construction via \ourengine and structured on-policy self-distillation. 
\Cref{alg:physopsd} details the latter, taking the paired
multi-level conditions constructed by \ourengine as input.

\newcommand{\mycommfont}[1]{\textcolor{headingpurple}{\textit{#1}}}

\begin{algorithm}[t]\small
\DontPrintSemicolon
\SetAlgoLined
\LinesNumbered
\SetCommentSty{mycommfont}

\caption{Structured OPSD in \ourmethod}
\label{alg:physopsd}

\KwIn{
Paired scene families $\mathcal{D}_{\mathrm{pair}}$
with conditions $\{\mathbf{c}_r^{(\ell)}\}_{r\in\{a,b\},\,\ell\in\{0,1,2\}}$;
pretrained parameters $\bm{\theta}_0$;
privilege distribution $P(\ell)$;
solver $\Phi$ and schedule $\{t_k\}_{k=0}^{K}$;
EMA decay $\beta$.
}
\KwOut{Trained student parameters $\bm{\theta}$.}

\tcp{Initialize student and EMA self-teacher weights}
$\bm{\theta} \gets \bm{\theta}_0$; \quad $\bar{\bm{\theta}} \gets \bm{\theta}_0$\;

\While{training budget remains}{

    $\mathcal{B} \gets \operatorname{SampleBatch}(\mathcal{D}_{\mathrm{pair}})$\;
    $\operatorname{ZeroGrad}(\bm{\theta})$\;

    \tcp{Paired on-policy self-distillation}
    \ForEach{$(W_a, W_b) \in \mathcal{B}$}{

        $\ell \sim P(\ell), \quad \bm{\epsilon} \sim \mathcal{N}(\mathbf{0}, \mathbf{I})$\;
        \tcp*{Shared privilege gap and initial noise}

        \ForEach{$r \in \{a, b\}$}{

            $\mathbf{c}_r^S \gets \mathbf{c}_r^{(\ell)}, \quad \mathbf{c}_r^T \gets \mathbf{c}_r^{(\ell+1)}$\;
            $\mathbf{x}_{t_K}^{S,r} \gets \bm{\epsilon}$\;

            \For{$k = K, K-1, \dots, 1$}{

                $\mathbf{u}_S^{r,k} \gets \mathbf{v}_{\bm{\theta}}(\mathbf{x}_{t_k}^{S,r}, t_k, \mathbf{c}_r^S)$\;
                $\mathbf{u}_T^{r,k} \gets \operatorname{sg}\left[ \mathbf{v}_{\bar{\bm{\theta}}}(\mathbf{x}_{t_k}^{S,r}, t_k, \mathbf{c}_r^T) \right]$\;

                $\mathcal{L}_{r,k} \gets \frac{1}{2|\mathcal{B}|K} \left\| \mathbf{u}_S^{r,k} - \mathbf{u}_T^{r,k} \right\|_2^2$\;
                $\operatorname{Backward}(\mathcal{L}_{r,k})$\;

                $\mathbf{x}_{t_{k-1}}^{S,r} \gets \operatorname{sg}\left[ \Phi(\mathbf{x}_{t_k}^{S,r}, t_k, t_{k-1}, \operatorname{sg}(\mathbf{u}_S^{r,k})) \right]$\;
            }
        }
    }

    \tcp{Optimizer and teacher EMA update}
    $\operatorname{ClipGradNorm}(\bm{\theta}, 1.0)$\;
    $\operatorname{AdamWStep}(\bm{\theta})$\;
    $\bar{\bm{\theta}} \gets \beta\bar{\bm{\theta}} + (1 - \beta)\bm{\theta}$\;
}

\Return{$\bm{\theta}$}\;
\end{algorithm}
\vspace{-0.6em}

\section{More Details of Experiments}
\label{app_sec:B}

\subsection{Baseline Implementations}
\label{app_sec:B_1}

\paragraph{Base Models.} 
We implement \ourmethod on two unified video generation backbones: \ding{182} OmniWeaving \citep{pan2026omniweaving}, which supports interleaved multimodal conditioning and free-form composition, and \ding{183} UniVideo \citep{wei2026univideo}, a dual-stream architecture unifying video understanding, generation, and editing. 
Both models natively accommodate text and visual inputs, providing the foundation for asymmetric student-teacher conditioning without architectural modifications.

\vspace{-1em}
\paragraph{Baselines.} 
As a post-training framework, our primary evaluation focuses on assessing relative capability gains against the respective base models. 
Furthermore, because physics-oriented post-training methods for video generation remain relatively scarce and are typically coupled with disparate backbones, we select and adapt three baselines to represent three prevailing post-training paradigms onto the same backbones:
\begin{itemize}
    \item \textbf{PhyT2V} \citep{xue2025phyt2v}: an iterative test-time refinement strategy. We preserve its physical chain-of-thought reasoning and prompt revision workflow, adapting its generator to our respective base models for direct paradigm comparison.
    \item \textbf{SFT} \citep{wang2026wisa}: standard supervised fine-tuning on physics-specific demonstrations. We fine-tune each base model on WISA-80K, which comprises approximately $80$K curated text-video pairs spanning 17 physical categories across dynamics, thermodynamics, and optics.
    \item \textbf{PhysHPO} \citep{chen2025hierarchical}: an offline hierarchical preference optimization framework. We adopt its original $21$K curated real-world videos from OpenVidHD \citep{nan2025openvid} as positive demonstrations, paired with base model rollouts as negative samples, tailored to OmniWeaving and UniVideo.
\end{itemize}

\vspace{-1em}
\paragraph{OPSD Variants.} 
For the privileged context analysis in \Cref{sec:analysis_privilege}, all variants are evaluated under an identical budget of $3$K samples. 
\ding{182} For \textit{w/ Refined Text}, we utilize the physical reasoning and prompt revision pipeline from PhyT2V on the real reference videos and their original prompts to synthesize enriched prompts with detailed physical descriptions, which serve as the privileged text context. 
\ding{183} For \textit{w/ Real Video}, the teacher receives both the text prompt and raw reference videos as privileged context, supplemented using PhysHPO's data selection pipeline to match the $3$K budget.

\subsection{Evaluation Benchmarks}
\label{app_sec:B_2}

\paragraph{VideoPhy-2.} 
VideoPhy-2 \citep{bansal2026videophy} evaluates action-centric physical commonsense across 200 action categories under the T2V setting. 
Following \citep{wang2026wisa}, we assess Semantic Adherence (SA) and Physical Commonsense (PC) by using GPT-5.5, reporting mean scores, the proportions of videos achieving $\text{SA} \ge 4$ and $\text{PC} \ge 4$, and the joint metric (All) satisfying both criteria simultaneously.

\vspace{-1em}
\paragraph{PhyGround.} 
PhyGround \citep{lin2026phyground} benchmarks text- and initial-frame-conditioned generation (TI2V) across 250 diverse scenarios spanning 13 physical laws in solid mechanics, fluid dynamics, and optics. 
Following its official benchmark protocol, we evaluate General plausibility (covering temporal coherence and object permanence) and law-specific Physics compliance, reporting their equal-weighted Overall score.

\vspace{-1em}
\paragraph{VBench.} 
We employ VBench 1.0 \citep{huang2024vbench} to benchmark general video generation capabilities under the T2V protocol. 
Following the official evaluation suite across 16 fine-grained dimensions, we report normalized Quality, Semantic, and Total scores using standard prompt sets.

\section{Exhibition Board}

\begin{figure*}[!t]
\centering
\includegraphics[width=\linewidth]{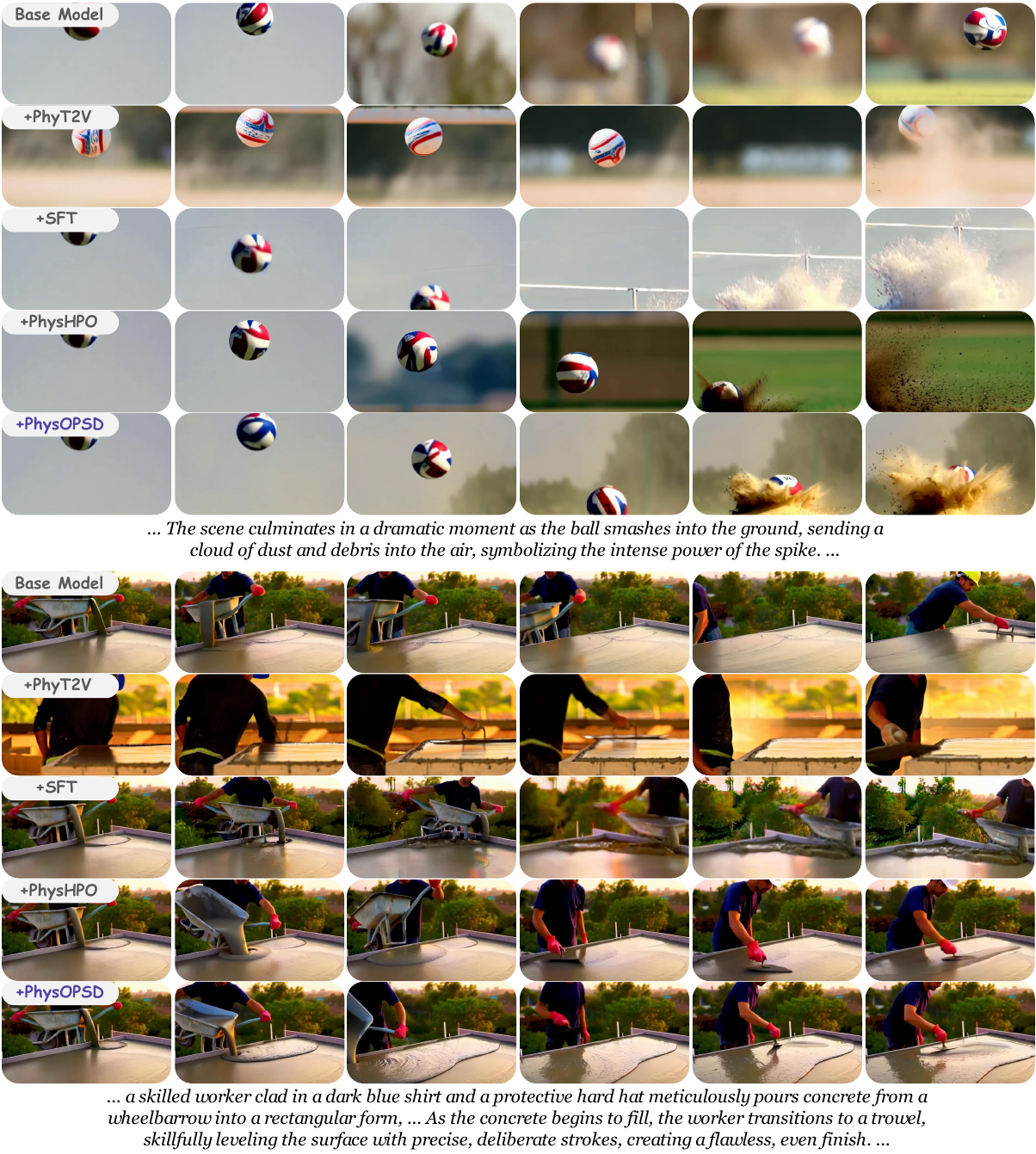}
\vspace{-1.6em}
\caption{Qualitative comparisons of \ourmethod with baselines.}
\label{app_fig:qualitative_baseline_1}
\end{figure*}

\begin{figure*}[!t]
\centering
\includegraphics[width=\linewidth]{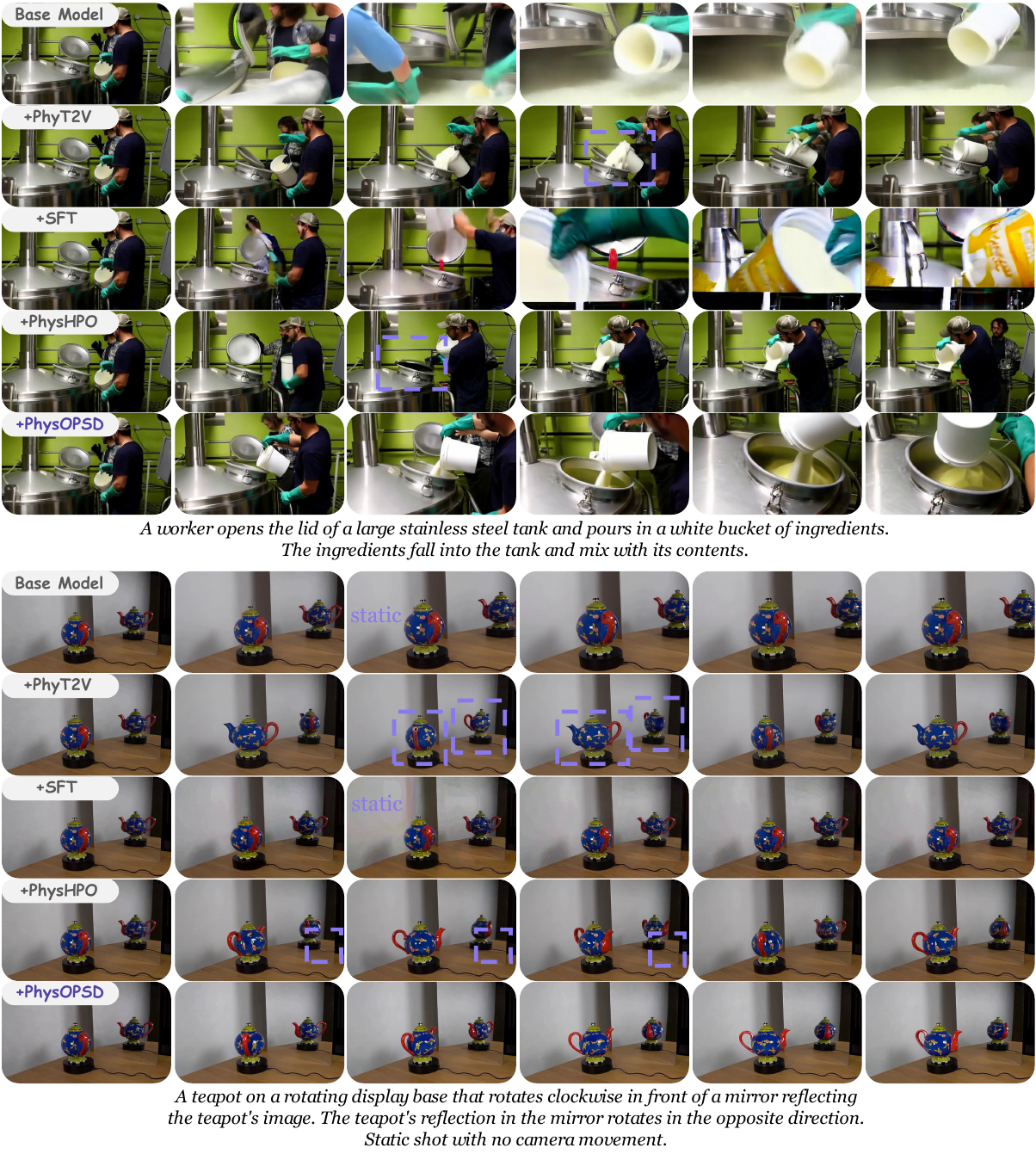}
\vspace{-1.6em}
\caption{Qualitative comparisons of \ourmethod with baselines.}
\label{app_fig:qualitative_baseline_2}
\end{figure*}

\begin{figure*}[!t]
\centering
\includegraphics[width=\linewidth]{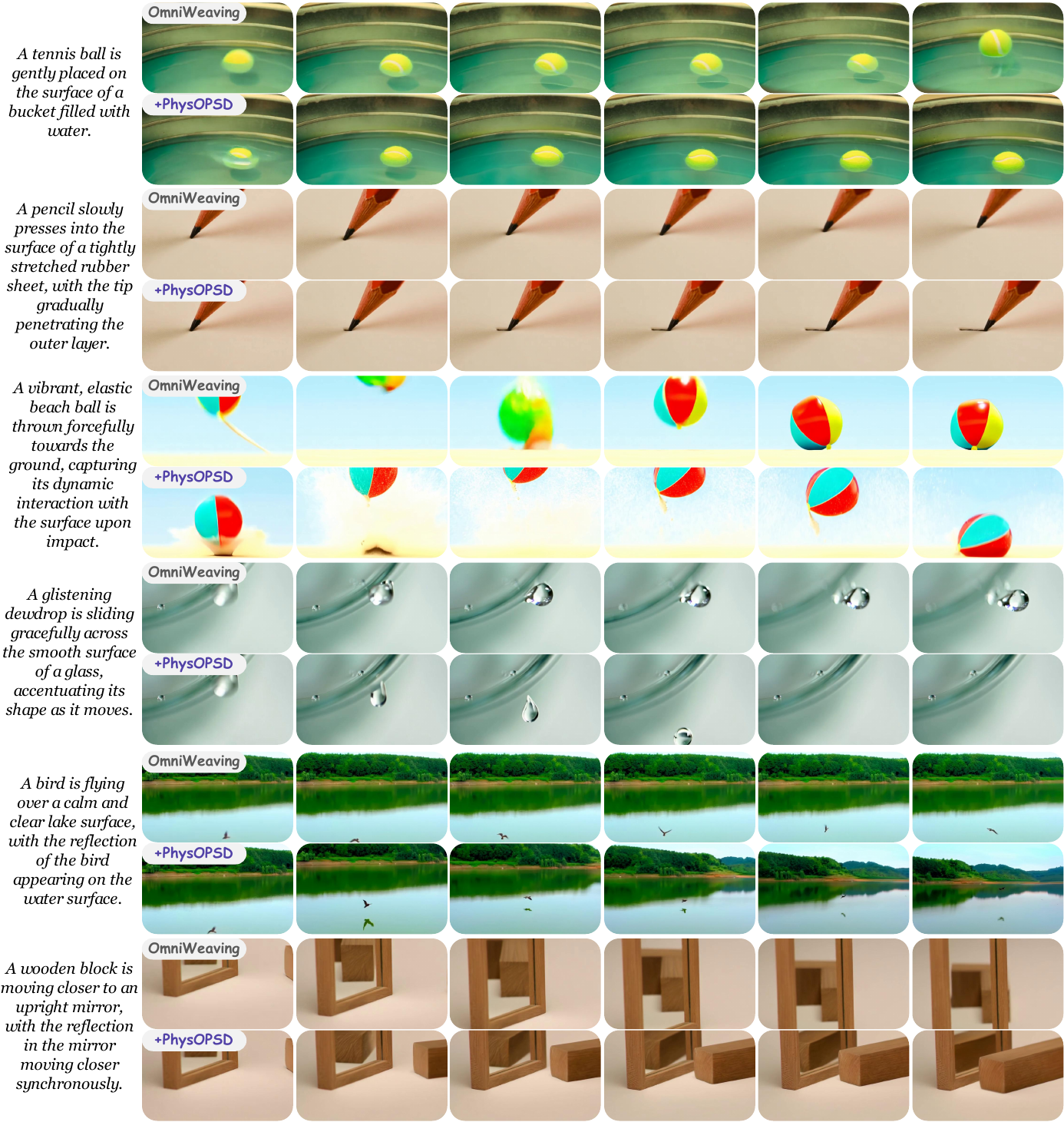}
\vspace{-1.6em}
\caption{Qualitative results of \ourmethod.}
\label{app_fig:qualitative_1}
\end{figure*}

\begin{figure*}[!t]
\centering
\includegraphics[width=\linewidth]{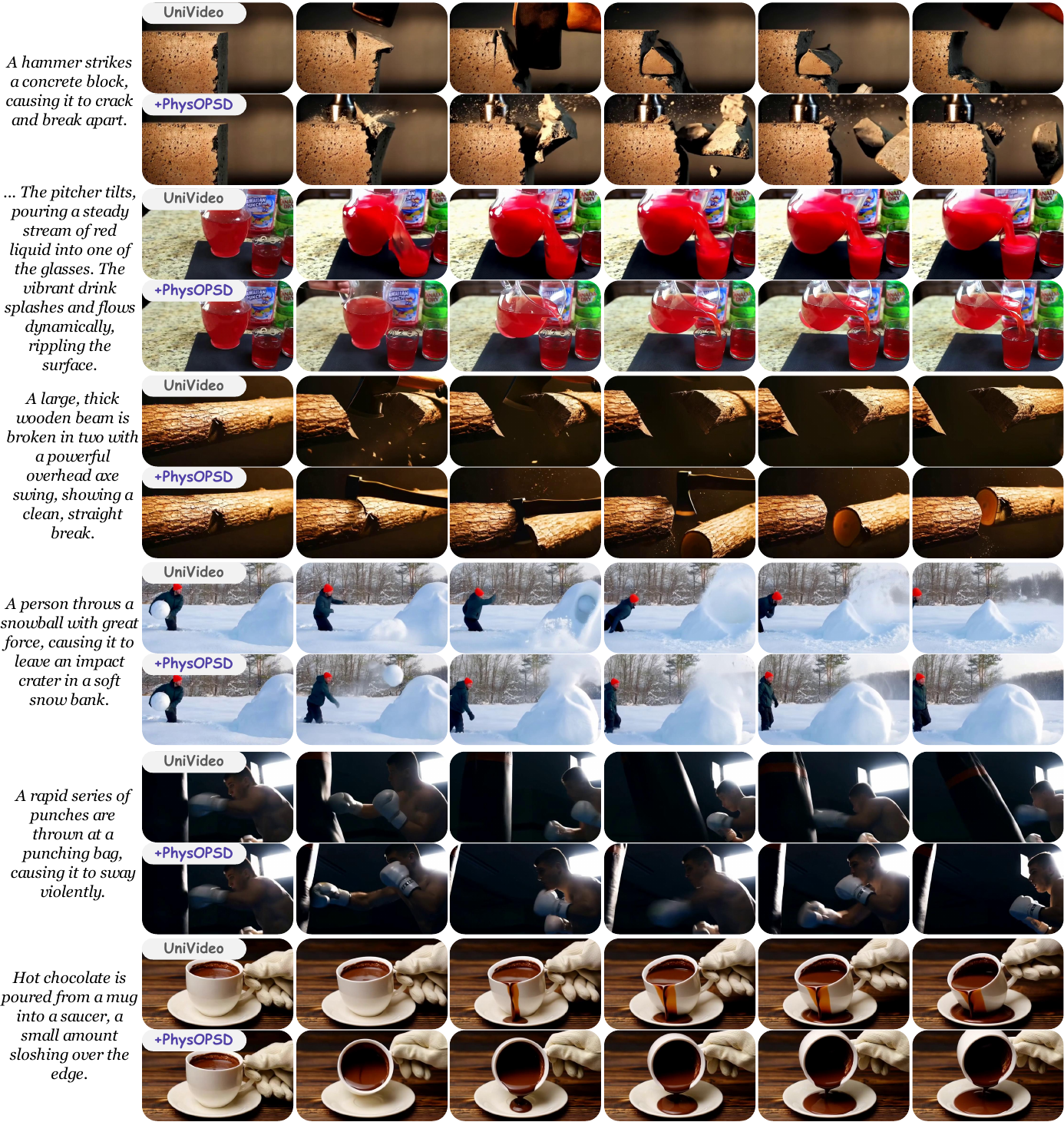}
\vspace{-1.6em}
\caption{Qualitative results of \ourmethod.}
\label{app_fig:qualitative_2}
\end{figure*}

We provide more comparison results of \ourmethod here in \Cref{app_fig:qualitative_baseline_1,app_fig:qualitative_baseline_2} with baselines: PhyT2V, SFT, and PhysHPO, along with more demonstrations in \Cref{app_fig:qualitative_1,app_fig:qualitative_2}.

\end{document}